%% file: cvmpaper_finalcopy.tex
\documentclass[10pt,twocolumn,letterpaper]{article}

\usepackage{cvm}
\usepackage{times}
\usepackage{epsfig}
\usepackage{graphicx}
\usepackage{amsmath}
\usepackage{amssymb}
\DeclareMathOperator*{\argmin}{argmin}

\usepackage{amsfonts}
\usepackage{verbatim}
\usepackage{multirow}
\usepackage{float}
\usepackage{subfigure}
\usepackage{algorithm}
\usepackage{algorithmic}
\usepackage[pagebackref=true,breaklinks=true,letterpaper=true,colorlinks,bookmarks=false]{hyperref}

\cvmfinalcopy

\def\cvmPaperID{47} 

\begin{document}

\title{Towards Uniform Point Distribution in Feature-preserving Point Cloud Filtering}

\author{Shuaijun Chen\\
Deakin University\\
Australia
\and
Jinxi Wang\\
Northwest A\&F University\\
Yangling, China
\and
Wei Pan\\
South China University of Technology\\
China
\and
Shang Gao\\
Deakin University\\
Australia
\and
Meili Wang\\
Northwest A\&F University\\
Yangling, China
\and
Xuequan Lu\\
Deakin University\\
Australia\\
{\small\url{http://www.xuequanlu.com}}
}

\maketitle

\begin{abstract}
As a popular representation of 3D data, point cloud may contain noise and need to be filtered before use. Existing point cloud filtering methods either cannot preserve sharp features or result in uneven point distribution in the filtered output. To address this problem, this paper introduces a point cloud filtering method that considers both point distribution and feature preservation during filtering. The key idea is to incorporate a repulsion term with a data term in energy minimization. The repulsion term is responsible for the point distribution, while the data term is to approximate the noisy surfaces while preserving the geometric features. This method is capable of handling models with fine-scale features and sharp features. Extensive experiments show that our method yields better results with a more uniform point distribution ($5.8\times10^{-5}$ Chamfer Distance on average) in seconds.
\end{abstract}

\input{paper/introduction}
\input{paper/relatedwork}
\input{paper/overview}
\input{paper/method}
\input{paper/results}
\input{paper/conclusion}

{\small
\bibliographystyle{cvm}
\bibliography{cvmbib}
}

\end{document}

%% file: paper/introduction.tex
\section{Introduction}

Researchers have made remarkable achievements in point cloud filtering in recent years. The newly proposed methods typically aim at maintaining the sharp features of the original point cloud while projecting the noisy points to underlying surfaces. The filtered point cloud data can then be used for upsampling \cite{huang2013edge_EAR_AWLOP}, surface reconstruction \cite{kazhdan2013screened_ScreenedPoisson,oztireli2009feature_RIMLS}, skeleton learning \cite{Lu_2018_skeleton_2,LU2019102792_skeleton_1} and computer animation \cite{Lu_animation_1,Pei_animation_2}, etc. 

The existing point cloud filtering methods can be divided into traditional and deep learning techniques. 
Among the traditional class, position-based methods \cite{lipman2007parameterization_LOP, huang2009consolidation_WLOP, preiner2014continuous_CLOP} obtain good smoothing results, while normal-based methods \cite{oztireli2009feature_RIMLS,lu2017gpf_GPF} achieve better effects in maintaining sharp edges of models (e.g., CAD models). Some of these methods incorporate repulsion terms to prevent the points from aggregating but still leave gaps near the edges of geometric features, which affects the reconstruction quality. 
Deep learning-based approaches \cite{rakotosaona2020pointcleannet_PointCleanNet,roveri2018pointpronets_PointProNets,zhang2020pointfilter_Pointfilter} require a number of noisy point clouds with ground-truth models for training and often achieve promising denoising performance through a proper number of iterations. These methods are usually based on local information, and lead to less even distribution in filtered results even in the presence of a ``repulsion'' loss term. It is difficult for these methods to handle unevenly distributed point clouds and sparsely sampled point clouds since the patch size is difficult to adjust automatically. Also, different patch sizes in the point cloud pose a significant challenge to the learning procedure.

The above analysis motivates us to produce filtered point clouds with the preservation of sharp features and a more uniform point distribution. 

In this paper, we propose a filtering method that preserves features well while making the points distribution more uniform.  
Specifically, given a noisy point cloud with normals as input, we first smooth the input normals using Bilateral Filtering \cite{huang2013edge_EAR_AWLOP}. 
Principal Component Analysis (PCA) \cite{hoppe1992surface} is used for the initial estimation of normals. 
Secondly, we update the point positions in a local manner by reformulating an objective function consisting of an edge-aware data term and a repulsion term inspired by  \cite{lu2020low_Lowrank,lu2017gpf_GPF}. The two terms account for preserving geometric features and point distribution, respectively. 
The uniformly distributed points with feature-preserving effects can be obtained through a few iterations. 
We conduct extensive experiments to compare our approach with various other approaches, including the position-based learning/traditional approaches and the normal-based learning/traditional approaches.
The results demonstrate that our method outperforms state-of-the-art methods in most cases, both in visualization and quantitative comparisons.

%% file: paper/relatedwork.tex
\section{Related Work}
\label{sec:relatedwork}
In this paper, we only review the most relevant work to our research, including traditional point cloud filtering and deep learning-based point cloud filtering.

\subsection{Traditional Point Cloud Filtering}

\textbf{Position-based methods.}
LOP was first proposed in \cite{lipman2007parameterization_LOP}. It is a parameterization-free method and does not rely on normal estimation. Besides fitting the original model, a density repulsion term was added to evenly control the point cloud distribution. 
WLOP \cite{huang2009consolidation_WLOP} provided a novel repulsion term to solve the problem that the original repulsion function in LOP dropped too fast when the support radius became larger. The filtered points were distributed more evenly under WLOP. EAR \cite{huang2013edge_EAR_AWLOP} added an anisotropic weighting function to WLOP to smooth the model while preserving sharp features. 
CLOP \cite{preiner2014continuous_CLOP} is another LOP-based approach. It redefined the data term as a continuous representation of a set of input points. 

Though only based on point positions, these approaches achieved fair smoothing results. Still, since they disregard normal information, these approaches tend to smear sharp features such as sharp edges and corners.

\textbf{Normal-based methods.}
FLOP \cite{liao2013efficient_FLOP} added normal information to the novel feature-preserving projection operator and preserved the features well. Meanwhile, a new Kernel Density Estimate (KDE)-based random sampling method was proposed for accelerating FLOP. 
MLS-based approaches \cite{levin1998approximation_MLS_1,levin2004mesh_MLS_2} have also been applied to point cloud filtering. They relied upon the assumption that the given set of points implicitly defined a surface. 
In \cite{alexa2003computing_crpssMLS}, the authors presented an algorithm that allocated a MLS local reference domain for each point that was most suitable for its adjacent points and further projected the points to the underlying plane.
This approach used the eigenvectors of a weighted covariance matrix to obtain the normals when the input point cloud had no normal information. 
APSS \cite{guennebaud2007algebraic_APSS}, RMLS \cite{rusu2007towards_RMLS}, and RIMLS \cite{oztireli2009feature_RIMLS} were implemented based on this, where RIMLS was based on robust local kernel regression and could obtain better results under the condition of higher noise. 
GPF \cite{lu2017gpf_GPF} incorporated normal information to Gaussian Mixture Model (GMM), which included two terms and performed well in preserving sharp features. 
A robust normal estimation method was proposed in \cite{lu2020low_Lowrank} for both point clouds and meshes with a low-rank matrix approximation algorithm, where an application of point cloud filtering was demonstrated. 
To keep the geometry features, \cite{liu2020feature} first filtered the normals by defining discrete operators on point clouds, and then present a bi-tensor voting scheme for the feature detection step. 

Inspired by image denoising, researchers have also investigated the nonlocal aspects of point cloud denoising. 
The nonlocal-based point cloud filtering methods \cite{deschaud2010point_nonlocal,digne2012similarity_nonlocal,zeng20193d_nonlocal,chen2019multi} often incorporated normal information and designed different similarity descriptions to update point positions in a nonlocal manner. 
Among them, \cite{deschaud2010point_nonlocal} proposed a similarity descriptor for point cloud patches based on MLS surfaces.
\cite{digne2012similarity_nonlocal} designed a height vector field to describe the difference between the neighborhood of the point with neighborhoods of other points on the surface.
Inspired by the low-dimensional manifold model, \cite{zeng20193d_nonlocal} extends it from image patches to point cloud surface patches, and thus serves as a similarity descriptor for nonlocal patches.
\cite{chen2019multi} presented a new multi-patch collaborative method that regards denoising as a low-rank matrix recovery problem. They define the given patch as a rotation-invariant height-map patch and denoise the points by imposing a graph constraint.

Filtering methods that rely on normal information usually yield good results, especially for point clouds with sharp features (e.g., CAD models).
However, these methods have a strong dependence on the quality of input normals, and a poor normal estimation may lead to worse filtering results.

Our proposed approach falls in the normal-based category. Inspired by GPF, we estimate normals of the input point cloud based on bilateral filtering \cite{huang2013edge_EAR_AWLOP} in order to get high-quality normal information. Note that if the input point cloud only contains positional information, PCA is used to compute the initial normals. The point positions are then updated in a local manner with the bilaterally filtered normals \cite{lu2020low_Lowrank}. We also add a repulsion term \cite{lu2020low_Lowrank} to ensure a more uniform distribution for filtered points.

\subsection{Deep Learning-based Point Cloud Filtering}
A variety of deep learning-based methods dealing with noisy point clouds have emerged \cite{liu2021treepartnet,erler2020points2surf,zhang2020pointfilter_Pointfilter,duan20193d_NPD,yu2018ec_EC-NET,rakotosaona2020pointcleannet_PointCleanNet,lu2020deep_dnp}.
In terms of point cloud filtering, PointProNets \cite{roveri2018pointpronets_PointProNets} introduced a novel generative neural network architecture that encoded geometric features in a local way and obtained an efficient underlying surface. However, the generated underlying surface was hard to fill the holes caused by input shapes. 
NPD \cite{duan20193d_NPD} redesigned the framework on the basis of PointNet \cite{qi2017pointnet_Pointnet} to estimate normals from noisy shapes and then projected the noisy points to the predicted reference planes. 
Another PointNet-inspired method is called Pointfilter \cite{zhang2020pointfilter_Pointfilter}. It started from points and learned the displacement between the predicted points and the raw input points. Moreover, this approach required normals only in the training phase. In the testing phase, only the point positions were taken as input to obtain filtered shapes with feature-preserving effects. 
EC-NET \cite{yu2018ec_EC-NET} presented an edge-aware network (similar to PU-NET \cite{yu2018pu_PU-NET}) for connecting edges of the original points. This method got promising results in retaining sharp edges in 3D shapes, but the training stage required manual labeling of the edges. 
Inspired by PCPNet \cite{guerrero2018pcpnet_PCPNET}, PointCleanNet \cite{rakotosaona2020pointcleannet_PointCleanNet} developed a data-driven method for both classifying outliers and reducing noise in raw point clouds. A novel feature-preserving normal estimation method was designed in \cite{lu2020deep_dnp} for point cloud filtering with preserving geometric features. 
Deep learning-based filtering methods usually yield good results with more automation and can often handle point clouds with high density. That is, low-density shapes as input may lead to poor filtering outcomes. Also, such methods require to ``see'' enough samples during training.

%% file: paper/overview.tex
\begin{figure*}[htbp]
\centering
\begin{minipage}[b]{0.95\linewidth}
{\label{}
\includegraphics[width=1\linewidth]{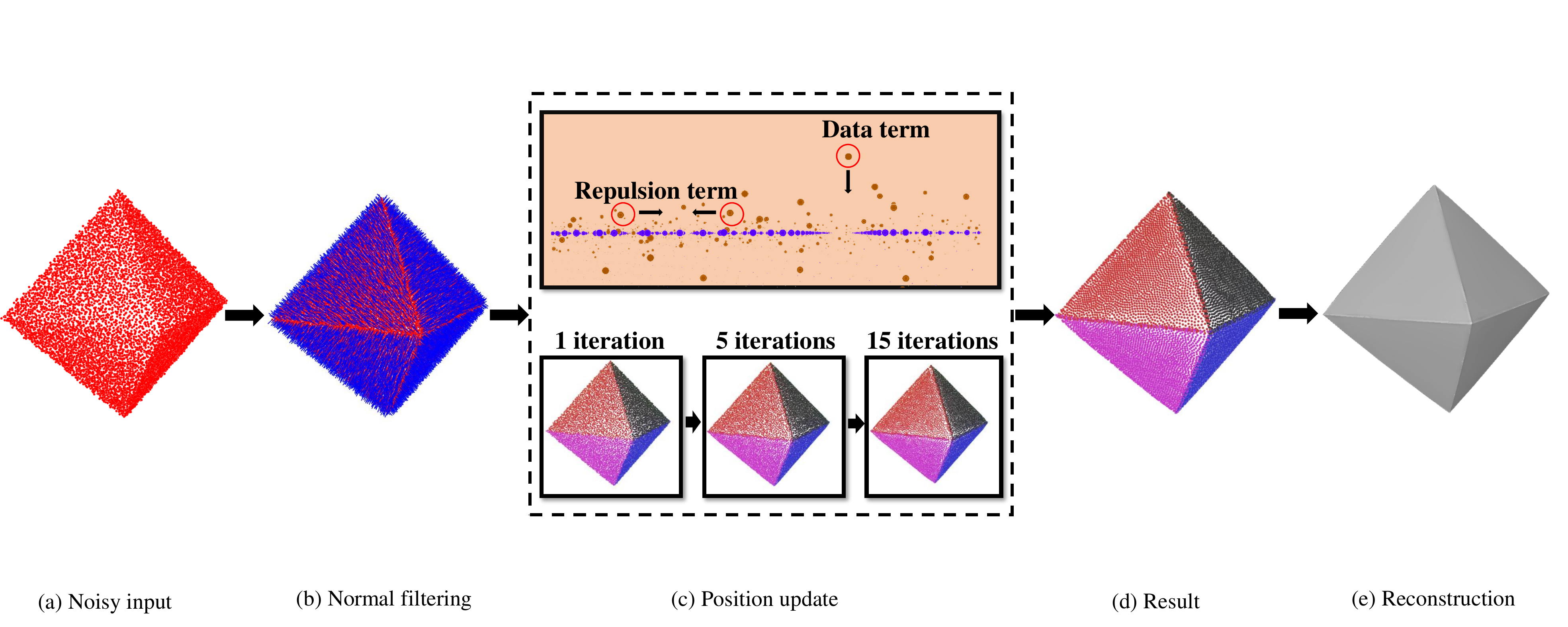}}
\end{minipage}
\caption{Overview of our approach. (a) Noisy input. Red color denotes points corrupted with noise. (b) Filtered normals. Blue lines denote filtered results of the initial normals. (c) Our position update method (considering a data term for feature preservation and a repulsion term for uniform distribution). Multiple iterations are performed to achieve a better filtered result. (d) The filtered point cloud. (e) The reconstructed mesh based on (d). }
\label{fig:overview}
\end{figure*}

%% file: paper/method.tex
\section{Method}
\label{sec:method}
Our approach consists of two phases. In phase one, we smooth the initial normals by Bilateral Filter (refer to \cite{huang2013edge_EAR_AWLOP} for more details) to ensure the quality of normals. In phase two, we update point positions with the smoothed normals to obtain a uniformly distributed point cloud with geometric features preserved. Figure \ref{fig:overview} shows an overview of the proposed approach. We will specifically explain the second phase in the following section.

\subsection{Position Update}
We first define a noisy input with $M$ points as $P={\{\mathbf{p}_i\}}_{i=1}^M, {\mathbf{p}_i} \in  {R^3}$ 
and $N={\{\mathbf{n}_i\}}_{i=1}^M, {\mathbf{n}_i} \in {R^3}$ as the corresponding filtered normals. To obtain local information from a given point $\mathbf{p}_i$, we define a local structure $\mathbf{s}_i$ for each point in the point cloud, consisting of the $k$ nearest points to the current point. We employ an edge-aware recovery algorithm \cite{lu2020low_Lowrank} to obtain filtered points by minimizing 
\begin{equation}\label{eq:pointminimization}
\begin{aligned}
\mathcal{D}(P,N)=\sum_{i} \sum_{j \in \mathbf{s}_{i}}\vert\left(\mathbf{p}_{i}-\mathbf{p}_{j}\right) \mathbf{n}_{j}^{T}\vert^{2}+\\
\vert\left(\mathbf{p}_{i}-\mathbf{p}_{j}\right) \mathbf{n}_{i}^{T}\vert^{2} ,
\end{aligned}
\end{equation}
where $\mathbf{p}_i$ denotes the point to be updated and $\mathbf{p}_j$ denotes the neighbor point in the corresponding set $\mathbf{s}_i$. Eq. \eqref{eq:pointminimization} essentially adjusts the angles between the tangential vector formed by $\mathbf{p}_i$ and $\mathbf{p}_j$ and the corresponding normal vectors $\mathbf{n}_i$, $\mathbf{n}_j$. 

\begin{figure}[!h]
\centering
\begin{minipage}[b]{0.88\linewidth}
\subfigure{\includegraphics[width=1\linewidth]{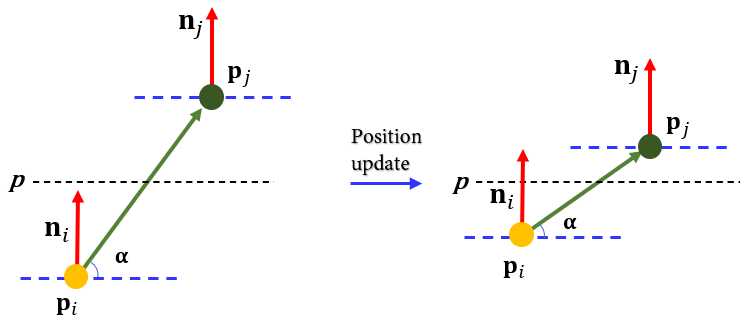}}
\label{fig:position update principle}
\end{minipage}
\caption{The left side represents the original points and the right side represents the updated points. $\mathbf{p}_i$ and $\mathbf{p}_j$ denote the current point and a neighboring point. $\mathbf{n}_i$, $\mathbf{n}_j$ denote normals of $\mathbf{p}_i$ and $\mathbf{p}_j$, respectively. Here a local plane surface is assumed. }
\label{fig:method}
\end{figure}

Figure \ref{fig:method} demonstrates how the points are updated on an assumed local plane by this edge-aware technique.
It can be seen that the quality of the filtered points depends heavily on the quality of the estimated normals. Our normals are generated by bilaterally filtering the original input normals, given its simplicity and effectiveness. 

\begin{figure}[!h]
\centering
\begin{minipage}[b]{0.9\linewidth}
\subfigure{\includegraphics[width=1\linewidth]{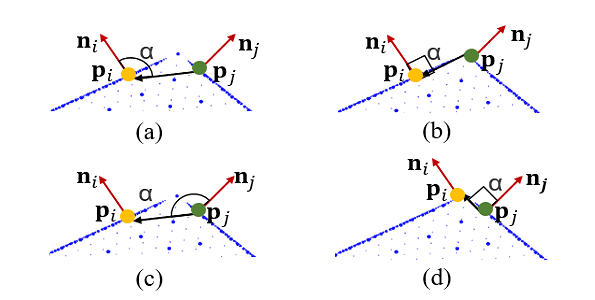}}
\end{minipage}
\caption{The movement of the filtered points around sharp edges. Blue points denote an underlying surface, yellow and green points indicate two neighboring points that need to be moved, respectively. (a-b) show the movement of $\mathbf{p}_j$ with fixed $\mathbf{p}_i$. (c-d) show the movement of $\mathbf{p}_i$ with fixed $\mathbf{p}_j$. This reveals that points are moving toward the sharp edges and concentrating there, leading to gaps around the sharp edges.}
\label{fig:method_distribution}
\end{figure}

\subsection{Repulsive Force}
From Figure \ref{fig:method_distribution}, it can be seen that the points would continually move towards the sharp edges at the position update step, thus inducing gaps near sharp edges.   
This is also demonstrated by \cite{lu2020low_Lowrank} that minimizing $\mathcal{D}(P,N)$ inevitably yields gaps near sharp edges, and the filtered points with obvious gaps might greatly impact following applications such as upsampling and surface reconstruction. Thus, we introduce $\mathcal{R}(P,N)$ \cite{lu2017gpf_GPF} to better control the distribution of points.
\begin{equation}\label{eq:repulsionterm}
    \begin{aligned}
    \mathcal{R}(P,N)=\sum_{i} \lambda_{i} \sum_{j \in s_{i}}^{M} \eta\left(r_{ij}\right) \theta\left(r_{ij}\right),
    \end{aligned}
\end{equation}
Eq. \eqref{eq:repulsionterm} obtains a repulsion force using both point coordinates and normals, where $r_{ij}=\left\|\left(\mathbf{p}_{i}-\mathbf{p}_{j}\right)-\left(\mathbf{p}_{i}-\mathbf{p}_{j}\right) \mathbf{n}_{j}^{T} \mathbf{n}_{j}\right\|$, the term $\eta(r)$ equals to $-r$, and the term $\theta(r)=e^{\left(-\mathbf{r}^{2} /(h / 2)^{2}\right)}$ denotes a smoothly decaying weight function.

\subsection{Minimization}
By combining Eq. \eqref{eq:pointminimization} and Eq. \eqref{eq:repulsionterm}, our final position update optimization becomes: 
\begin{equation}\label{eq:allminimization}
    \begin{aligned}
    \argmin_{P}\mathcal{D}(P,N)+\mathcal{R}(P,N)
    \end{aligned}
\end{equation}

The gradient descent method is employed to minimize Eq. \eqref{eq:allminimization} and obtain the updated point $\mathbf{p}_i'$.
The partial derivative of Eq. \eqref{eq:allminimization} with respect to $\mathbf{p}_i$ is:
\begin{equation}
    \begin{split}
\frac{\partial}{{\partial}{\mathbf{p}_i}}=\sum_{j \in \mathbf{s}_{i}}\frac{\left(\mathbf{n}_{j} \mathbf{p}_{i}^{T}-\mathbf{n}_{j}\mathbf{p}_{j}^{T}\right)\left(\mathbf{p}_{i}\mathbf{n}_{j}^{T}-\mathbf{p}_{j} {\mathbf{n}_{j}}^{T}\right)}{{\partial}\mathbf{p}_i}+\\
           \frac{\lambda_{i}\beta_{ij}\left(\mathbf{p}_{i}-\mathbf{p}_{j}\right) (\mathbf{I}-\mathbf{n}_{j}^{T} \mathbf{n}_{j})}
    {{\partial}\mathbf{p}_i},
    \end{split}
\end{equation}
where $\beta_{ij}$ denotes $\frac{\theta\left(r_{ij}\right)}{r_{ij}}\left|\frac{\partial \eta\left(r_{ij}\right)}{\partial r}\right|$, and $\mathbf{I}$ is a $3\times3$ identity matrix. 

The updated point $\mathbf{p}_i'$ can be calculated by:
\begin{equation}\label{eq:pointupdate}
    \begin{aligned}
    \mathbf{p}_{i}^{\prime}=\mathbf{p}_{i}+\gamma_{i} \sum_{j \in s_{i}}\left(\mathbf{p}_{j}-\mathbf{p}_{i}\right)\left(\mathbf{n}_{j}^{T} \mathbf{n}_{j}+\mathbf{n}_{i}^{T} \mathbf{n}_{i}\right)+\\
    \mu \frac{\sum_{j \in s_{i}} w_{j} \beta_{i j}\left(\mathbf{p}_{i}-\mathbf{p}_{j}\right) (\mathbf{I}-\mathbf{n}_{j}^{T} \mathbf{n}_{j})}{\sum_{j \in s_{i}} w_{j} \beta_{i j}},
    \end{aligned}
\end{equation}
where $\gamma_{i}$ is set to $\frac{1}{3\left|s_{i}\right|}$ according to \cite{lu2020low_Lowrank}, $w_{j}$ denotes $1+\sum_{j \in s_{i}} \theta \left( \left \|\mathbf{p}_i-\mathbf{p}_j\right\|\right)$, and $\mu$ is a parameter which aims at controlling the relative magnitude of the repulsive force.

\subsection{Algorithm}
The proposed method is described in Algorithm \ref{alg:Framwork}. We first filter the normals using bilateral filtering. By feeding the filtered normals and raw point positions into Algorithm \ref{alg:Framwork}, we can obtain the updated point positions. Depending on the point number of each model and the noise level, we choose different $k$ to generate the local patches and perform several iterations accordingly. Section \ref{sec:results} provides the filtered results of different models. Table \ref{table:parameter_settings} lists our parameter settings.

\begin{algorithm}
    \begin{algorithmic}
        \STATE \textbf{Input:} Noisy point set $P$, corresponding filtered normals $N$.
        \STATE \textbf{Output:} Uniformly distributed set of filtered points $P'$.
        \STATE \textbf{Initialize:} iteration $t$, repulsion term $\mu$, local patch $\mathbf{s}_i$
        \FOR{Each iteration}
        \FOR{Each point $\mathbf{p}_i$}
            \STATE construct a local patch $\mathbf{s}_i$;
            \STATE update point position via Eq. \eqref{eq:pointupdate};
        \ENDFOR
        \ENDFOR
    \end{algorithmic}
    \caption{Towards uniform point distribution in feature-preserving point cloud filtering}\label{alg:Framwork}
\end{algorithm}

%% file: paper/results.tex
\section{Experimental Results}
\label{sec:results}

The proposed method is implemented in Visual Studio 2017 and runs on a PC equipped with i9-9750h and RTX2070. Most examples in this paper are executed in 7 seconds. The most time-demanding is the object from Figure \ref{fig:Realscan} that takes 28 seconds.

\subsection{Parameter Setting}
The parameters include the local neighborhood size $k$, the coefficient of repulsion force $\mu$, and the number of iterations $t$. 
Considering that the number of points affects the range of neighbors significantly, in order to find the appropriate $k$ neighbors for different models, here we determine the size of $k$ in the range [15, 45] ($k$ = 30 by default) according to the point number of each model. 
To make the distribution of points more even while preserving the features, we use the parameter $\mu$ to balance the magnitude of the repulsive force among points and $t$ to control the number of iterations. 
For the models with sharp edges, we set a relatively large $\mu$ with a low number of iterations, specifically $\mu=0.3$, $t=10$ or $t=5$. 
For models with smooth surfaces (e.g., non-CAD model), we use a smaller value of $\mu$ and a higher number of iterations $t$ with $\mu=0.1$, $t=30$. 
Table \ref{table:parameter_settings} gives all the parameters of the models used in the experiments.

\begin{table}[!h]
\centering
\caption{Parameter settings for different models. }
\begin{tabular}{l l l l l l}
\hline
Parameters     & $k$  &$\mu$ &$t$\\
\hline
Figure \ref{fig:bunnyhi}      & 30    & 0.3    & 5    \\

Figure \ref{fig:rockerman}  & 30    & 0.3    & 5    \\

Figure \ref{fig:Icosahedron}   & 30    & 0.3    & 5    \\

Figure \ref{fig:Dodecahedron}    & 30    & 0.3    & 3    \\

Figure \ref{fig:kitten}    & 30    & 0.3    & 5    \\

Figure \ref{fig:Nefertiti} & 30    & 0.3    & 5    \\

Figure \ref{fig:BuddhaStele}& 30    & 0.3    & 10    \\

Figure \ref{fig:Realscan} & 30    & 0.3    & 5    \\

Figure \ref{fig:david} & 30    & 0.3    & 5    \\
\hline
\end{tabular}
\label{table:parameter_settings}
\end{table}

\subsection{Compared Approaches}
The proposed method is compared with the state-of-the-art techniques which include the non-deep learning position-based method CLOP \cite{preiner2014continuous_CLOP}, non-deep learning normal-based methods GPF \cite{lu2017gpf_GPF} and RIMLS \cite{oztireli2009feature_RIMLS}, and deep learning-based methods TotalDenoising (TD) \cite{Hermosilla_2019_ICCV_TotalDenoising}, PointCleanNet (PCN) \cite{rakotosaona2020pointcleannet_PointCleanNet} and Pointfilter (PF) \cite{zhang2020pointfilter_Pointfilter}.
We employ the following rules for a fair comparison: (a) We first normalize and centralize the noisy input. (b) As GPF and RIMLS all require high-quality normals, we adopt the same Bilateral Filter \cite{huang2013edge_EAR_AWLOP} to obtain the same input normals for each model. (c) We try our best to tune the main parameters of each method to produce their final visual results. (d) For the deep learning-based methods, we use the results of the 6\textsuperscript{th} iteration for TD and iterate three times for both PCN and PF. (e) For visual comparison, we use EAR \cite{huang2013edge_EAR_AWLOP} for upsampling and achieve a similar number of upsampling points with the same model. As for surface reconstruction, we adopt the same parameters for the same model. 

\subsection{Evaluation Metrics}
Before discussing the visual results, we introduce two common evaluation metrics for analyzing the performance quantitatively. Suppose the ground-truth point cloud and the filtered point cloud are respectively defined as: 
${S_{1}}=\left\{\mathbf{x}_{i}\right\}_{i=1}^{|S_{1}|}, {S_{2}}=\left\{\mathbf{y}_{i}\right\}_{i=1}^{|S_{2}|}$. Notice that the number of ground-truth points $|S_{1}|$ and filtered points $|S_{2}|$ may be slightly different.

\begin{itemize}
    \item[1)]
    \textit{Chamfer Distance:}
    \begin{equation}\label{error_chamferdistance}
        \begin{aligned}
        \begin{split}
            e_{\mathrm{CD}}\left(S_{1}, S_{2}\right)=\frac{1}{|S_{1}|} \sum_{x \in S_{1}} \min _{y \in S_{2}}\|x-y\|_{2}^{2}+\\
            \frac{1}{|S_{2}|} \sum_{y \in S_{2}} \min _{x \in S_{1}}\|y-x\|_{2}^{2},        
        \end{split}
        \end{aligned}
    \end{equation}
    \item[2)]
    \textit{Mean Square Error:}
    \begin{equation}\label{error_mse}
        \begin{aligned}
        \begin{split}
            e_{\mathrm{MSE}}(S_{1}, S_{2}) = \frac{1}{|S_{1}|} \frac{1}{|NN(y)|} \sum_{x\in S_{1}}\\
            \sum_{y\in NN(y)} \|x-y\|_{2}^{2},            
        \end{split}
        \end{aligned}
    \end{equation}
where $NN(y)$ denotes the nearest neighbors in $S_1$ for point $y$ in $S_2$. Here we set $|NN(y)|=10$ similar to \cite{zhang2020pointfilter_Pointfilter}, which means we search 10 nearest neighbors for each point $y$ in the predicted point set $S_2$.
\end{itemize}

\begin{figure*}[htb]
\centering
\begin{minipage}[b]{0.95\linewidth}
{\label{}
\includegraphics[width=1\linewidth]{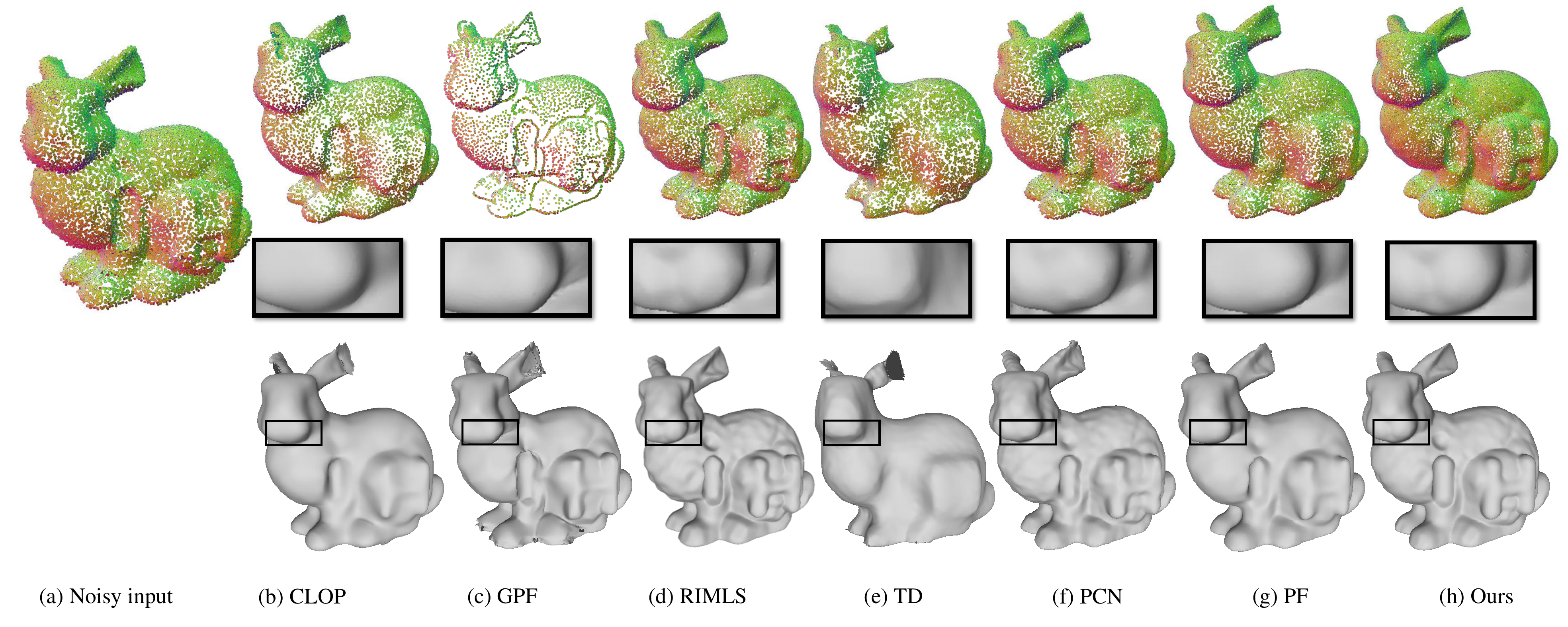}}
\end{minipage}
\caption{Results on the Bunnyhi model corrupted with 0.5\% synthetic noise. The second row gives the surface reconstruction results. }
\label{fig:bunnyhi}
\end{figure*}

\begin{figure*}[h]
\centering
\begin{minipage}[b]{0.95\linewidth}
{\label{}
\includegraphics[width=1\linewidth]{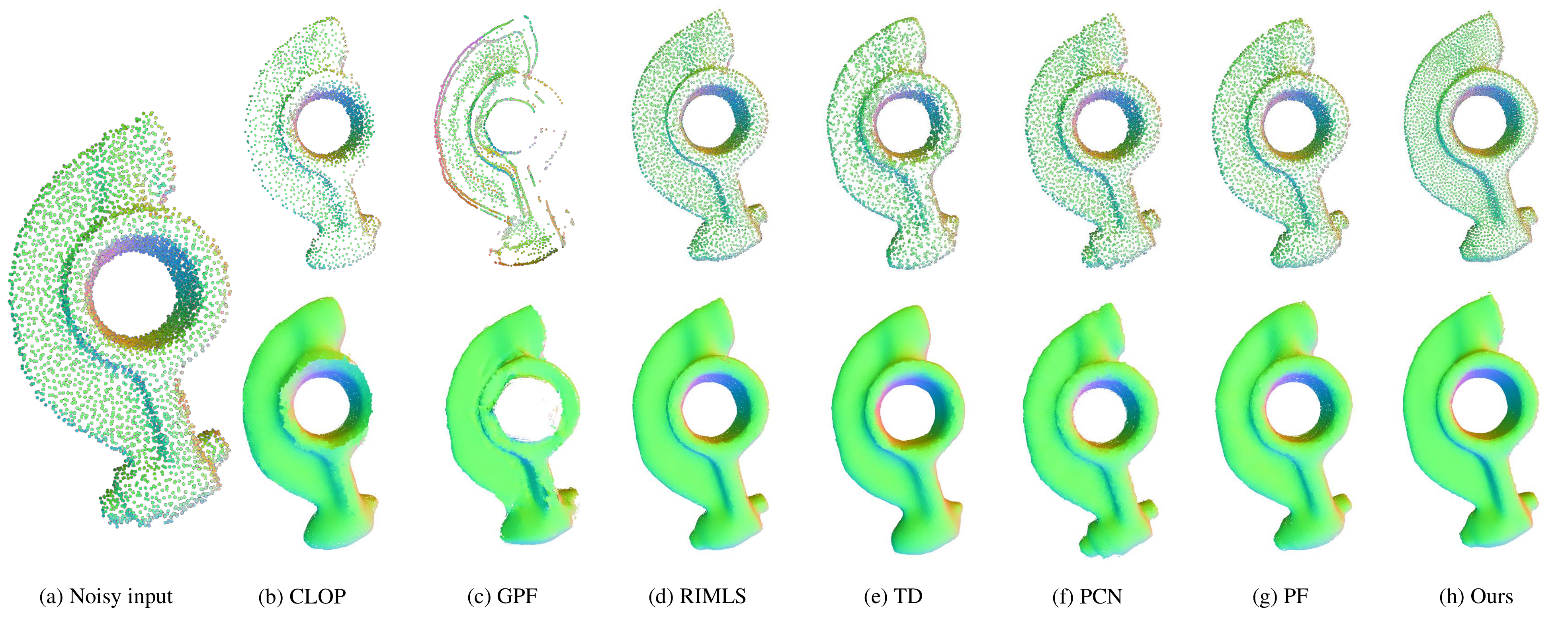}}
\end{minipage}
\caption{Results on Rockerman corrupted with 0.5\% synthetic noise. The second row gives the corresponding upsampling results. }
\label{fig:rockerman}
\end{figure*}

\begin{figure*}[!h]
\centering
\begin{minipage}[b]{0.95\linewidth}
{\label{}
\includegraphics[width=1\linewidth]{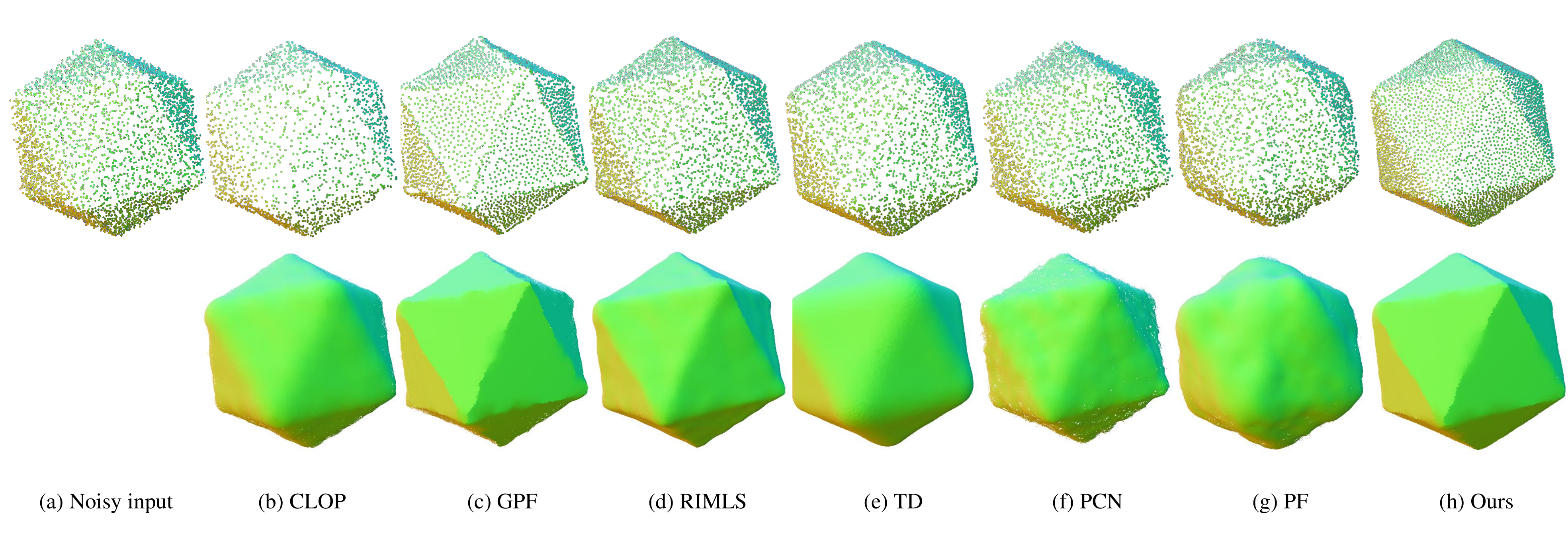}}
\end{minipage}
\caption{Results on Icosahedron corrupted with 1.0\% synthetic noise. The second row gives the corresponding upsampling results. }
\label{fig:Icosahedron}
\end{figure*}

\begin{figure*}[h]
\centering
\begin{minipage}[b]{0.95\linewidth}
{\label{}
\includegraphics[width=1\linewidth]{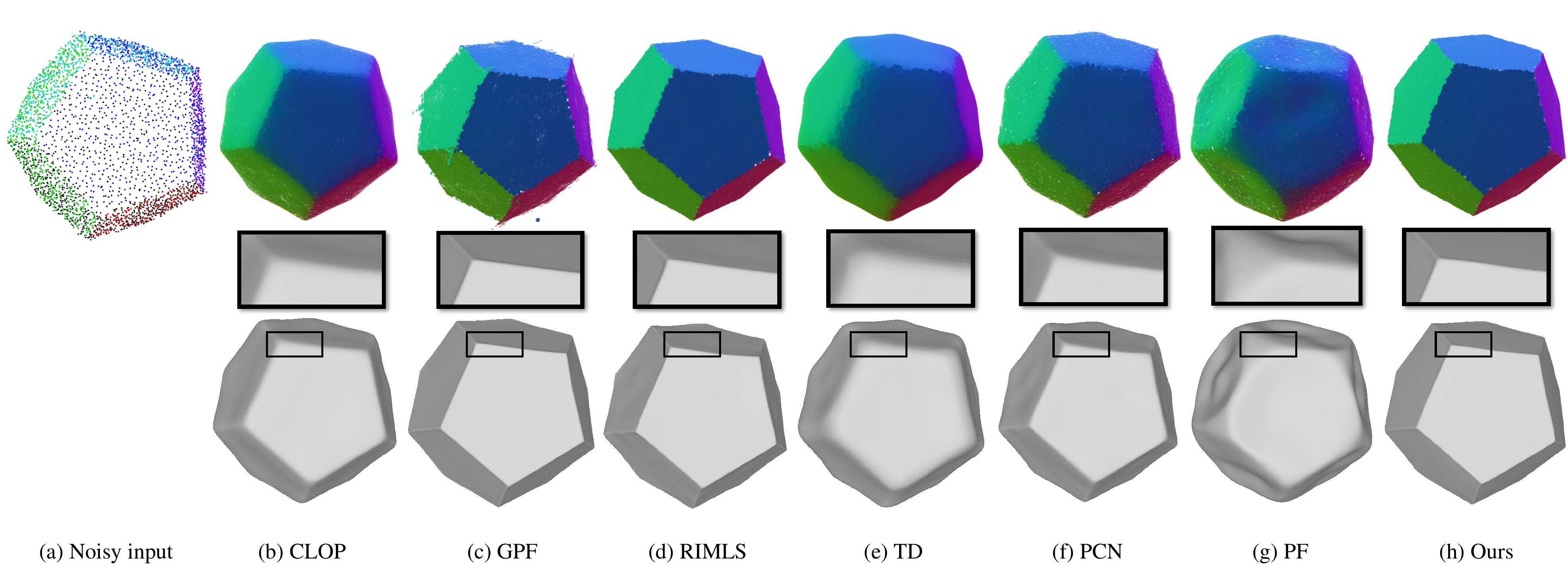}}
\end{minipage}
\caption{Results on Dodecahedron corrupted with 0.5\% synthetic noise. The first row gives the corresponding upsampling results and the reconstruction meshes are shown at the bottom. }
\label{fig:Dodecahedron}
\end{figure*}

\begin{figure*}[h]
\centering
\begin{minipage}[b]{0.95\linewidth}
{\label{}
\includegraphics[width=1\linewidth]{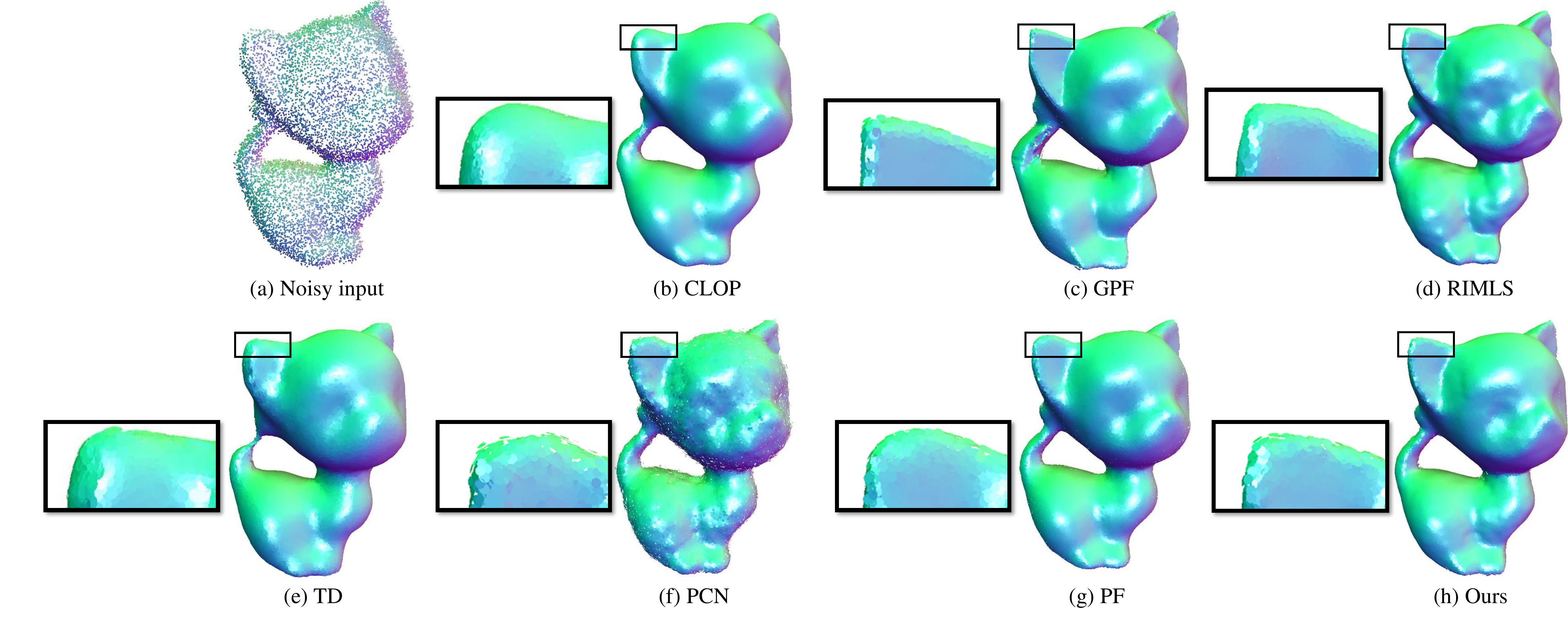}}
\end{minipage}
\caption{Results on kitten corrupted with 1.0\% synthetic noise. upsampling is included. }
\label{fig:kitten}
\end{figure*}

\begin{figure*}[h]
\centering
\begin{minipage}[b]{0.95\linewidth}
{\label{}
\includegraphics[width=1\linewidth]{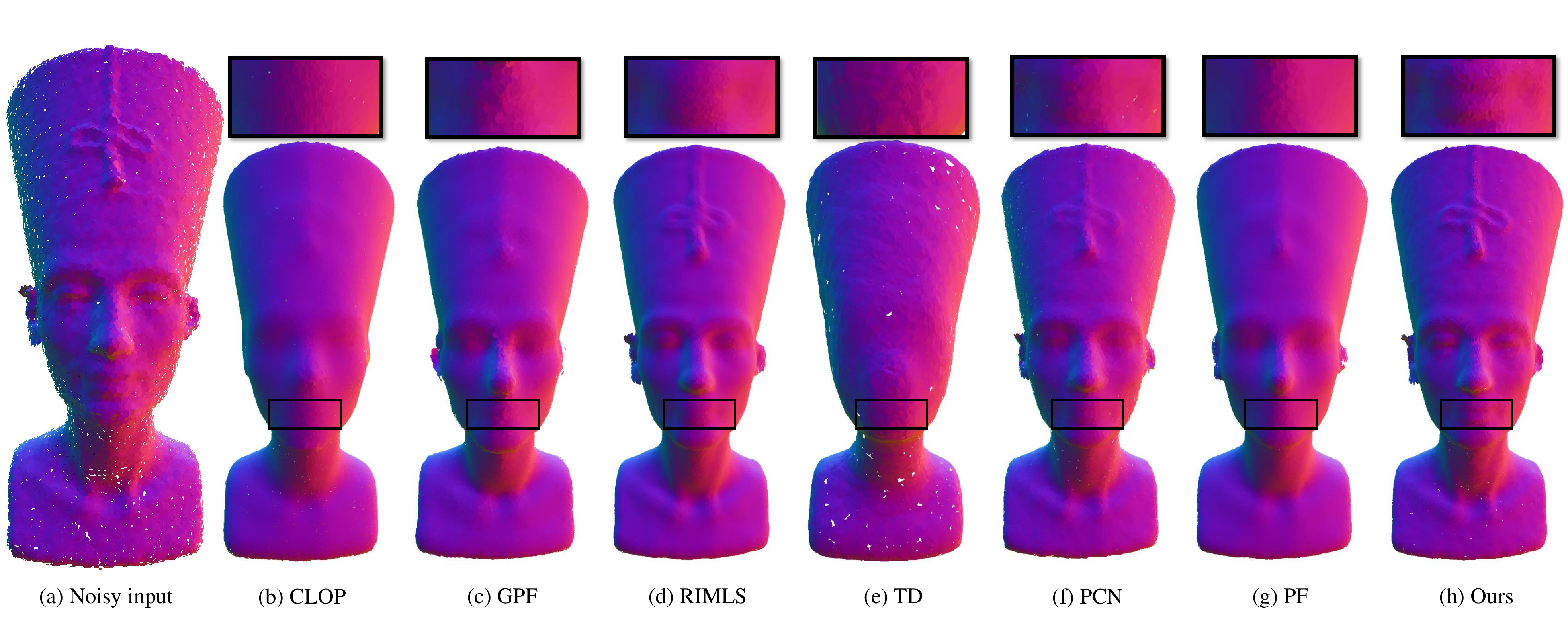}}
\end{minipage}
\caption{Results on one raw scanned model named Nefertiti. Upsampling is included. }
\label{fig:Nefertiti}
\end{figure*}

\begin{figure*}[h]
\centering
\begin{minipage}[b]{0.95\linewidth}
{\label{}
\includegraphics[width=1\linewidth]{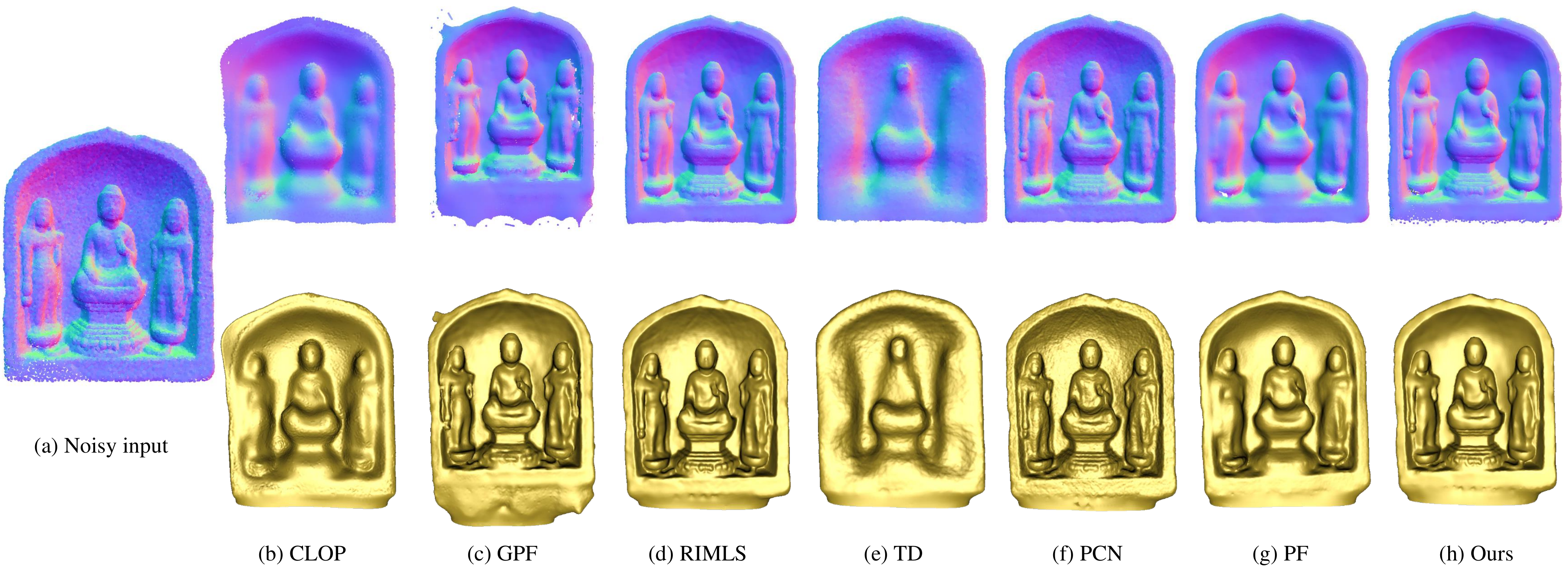}}
\end{minipage}
\caption{Results on the raw scanned model named BuddhaStele. The first row gives the corresponding upsampling results and the reconstruction meshes are shown at the bottom. }
\label{fig:BuddhaStele}
\end{figure*}

\begin{figure*}[h]
\centering
\begin{minipage}[b]{0.95\linewidth}
{\label{}
\includegraphics[width=1\linewidth]{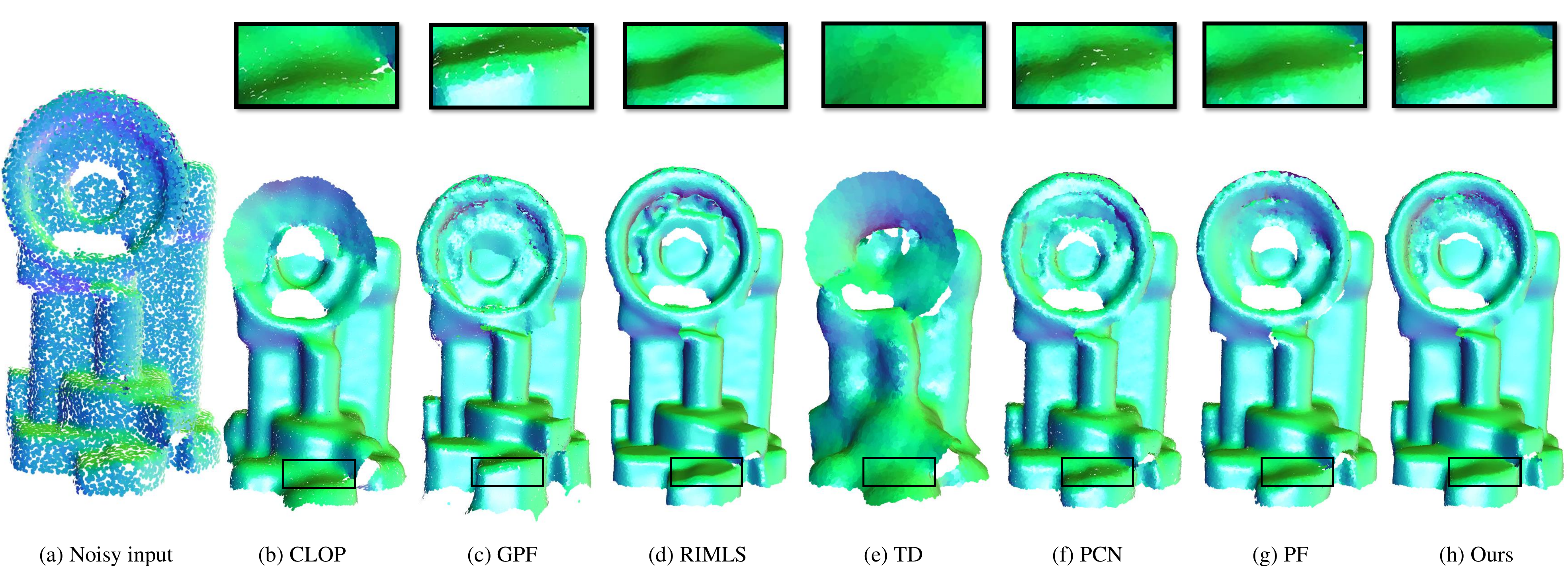}}
\end{minipage}
\caption{Results on the raw scanned model named Realscan. Upsampling is included. }
\label{fig:Realscan}
\end{figure*}

\begin{figure*}[h]
\centering
\begin{minipage}[b]{0.95\linewidth}
{\label{}
\includegraphics[width=1\linewidth]{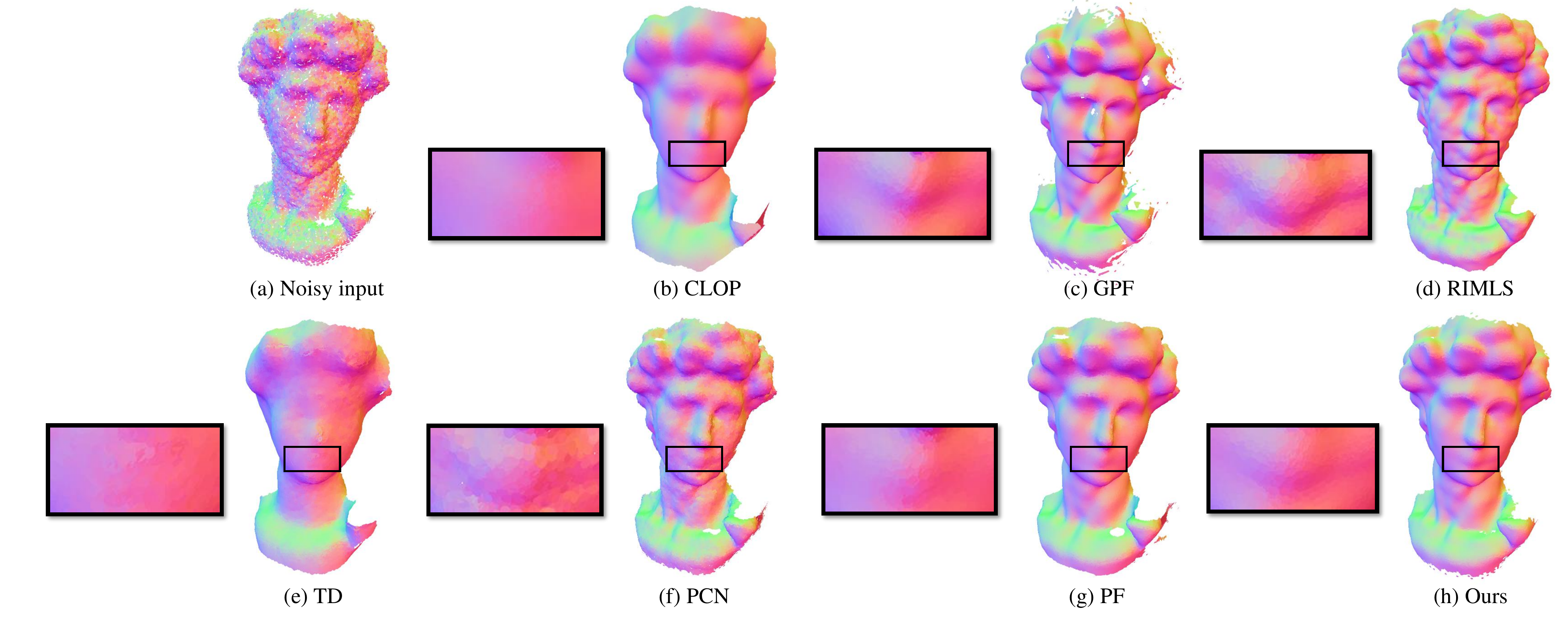}}
\end{minipage}
\caption{Results on the raw scanned model named David. Upsampling is included. }
\label{fig:david}
\end{figure*}

\subsection{Visual Comparisons}
\textbf{Point clouds with synthetic noise.} 
To show the denoising effect of our method, we conduct experiments on models with synthetic noise, which are Gaussian noise at levels of 0.5\% and 1.0\%, respectively. Compared to other state-of-the-art methods, our visual results outperform them in terms of both smoothing and feature-preserving aspects. The results benefit from the fact that the position update considers normal information makes the filtered points distribute more evenly. 

Meanwhile, we also observe the traits of other methods in the experiments. 
CLOP always obtains good results in terms of smoothing. However, since it is a position-based method, it may blur sharp features. 
While GPF adds a gap-filling step after projecting the points onto the underlying surface, it is still difficult to maintain a uniform distribution, especially when points are near sharp edges. This method may also make those models with less sharp features abnormally sharp.   
RIMLS yields promising results in both noise removal and feature preservation. Still its filtered points are often unevenly distributed, which affects the performance in following applications such as upsampling and surface reconstruction. 

The learning-based method TD also yields good smoothing results, but it does not seem to maintain the fine features of the model well.  
PCN typically produces less sharp features, and PCN can hardly obtain good smoothing effects under a relatively high level of noise.   
PF does not need normal information at the test stage, but it can still achieve a good feature-preserving effect while denoising. However, when the noisy points are sparse, this method cannot extract the information from the sparse point cloud, leading to distortion of the filtered points.

Since the normal information is taken into account for our method, it can keep the sharp features well. Importantly, the better uniform point distribution makes it stand out in point cloud filtering and following applications like upsampling and surface reconstruction.  

From the first row of Figures \ref{fig:bunnyhi}, \ref{fig:rockerman} and \ref{fig:Icosahedron}, it can be easily seen that our method obtains the most uniform point distribution. 
Figures \ref{fig:rockerman}, \ref{fig:Icosahedron} and \ref{fig:kitten} give the upsampling results of the three different models after filtering. As seen in the second rows of Figure \ref{fig:rockerman} and Figure \ref{fig:Icosahedron}, the sharp edges are maintained well during denoising. 
The enlarged box in Figure \ref{fig:kitten} also shows the effect of our filtering method, where the shape of the kitten's ears is maintained quite well. 
The results of surface reconstruction are presented in Figures \ref{fig:bunnyhi} and \ref{fig:Dodecahedron}. As can be seen from the enlarged box, we maintain the bunny's mouth and nose features in Figure \ref{fig:bunnyhi} very well. Figure \ref{fig:Dodecahedron} also shows the filtered results of our method on a simple geometric model with sharp edges. Our method is the best in terms of maintaining details and sharp edges.

\textbf{Point clouds with raw scan noise.}
In addition to the synthetic noise, we also perform experiments on raw scanned point clouds. The results of our method and existing methods on different raw point clouds are given in Figures \ref{fig:Nefertiti}, \ref{fig:BuddhaStele}, \ref{fig:Realscan} and \ref{fig:david}.
The filtered results of ours and other approaches given in Figure \ref{fig:Nefertiti} show that our method performs better in terms of smoothing and preserving details. As can be seen from the enlarged box, most methods make the mouth of the model blurry or disappear after denoising. Note that although the model we use here has the same shape as PF \cite{zhang2020pointfilter_Pointfilter}, the filtered results may be different since our sampling points are sparser than theirs. 

Figure \ref{fig:BuddhaStele} shows the filtered results on a raw scanned model named BuddhaStele. The first row gives the results after upsampling, and the second row provides the results of surface reconstruction using Screened Poisson \cite{kazhdan2013screened_ScreenedPoisson}. From the details such as the stairs in the model, it can be seen that our method still outperforms the other methods. 
In Figure \ref{fig:Realscan}, our method maintains the sharp edges well on this model. As seen in the zoom-in box, other state-of-the-art methods either distort the sharp edges or smooth them out. 
Figure \ref{fig:david} shows filtered results on a raw scanned model named David. Our method preserves feature better during filtering. Seeing details from the zoom-in box, our approach maintains the facial features better than others. 

\begin{table*}
    \centering
    \caption{ Quantitative evaluation results of the compared methods and our method on the synthetic point clouds in Figures \ref{fig:bunnyhi}, \ref{fig:rockerman}, \ref{fig:Icosahedron}, \ref{fig:Dodecahedron} and \ref{fig:kitten}. Note that * represents deep learning methods. Chamfer Distance ($\times10^{-5}$) is used here. The best method for each model is highlighted. }
    \label{table:errors_evaluation_chamfer_syn}
    \begin{tabular}{l c c c c c c}
        \hline
        Methods& {Figure \ref{fig:bunnyhi}}& {Figure \ref{fig:rockerman}}& {Figure \ref{fig:Icosahedron}}& {Figure \ref{fig:Dodecahedron}} & {Figure \ref{fig:kitten}}& Avg.\\ 
        \hline
        CLOP \cite{preiner2014continuous_CLOP}            & 7.84          & 25.35           & 26.46 & 23.83          & 6.73          & 16.70\\
        GPF \cite{lu2017gpf_GPF}                         & 16.19         & 31.85          & 21.52         & 18.35         & 15.54          & 17.58\\
        RIMLS \cite{oztireli2009feature_RIMLS}             & 3.72         & \textbf{5.22}           & 15.70         & 10.98 & 4.16          & 7.12\\
        TD* \cite{Hermosilla_2019_ICCV_TotalDenoising}& 23.88 & 13.20  & 24.43          & 19.22 & 11.86  & 16.15\\
        PCN* \cite{rakotosaona2020pointcleannet_PointCleanNet} & 4.76 & 6.38  & 29.87          & 14.96 & 6.24  & 11.18\\
        PF* \cite{zhang2020pointfilter_Pointfilter}          & 4.01 & 6.63  & 33.38 & 30.60 & 3.68  & 14.93\\
        \hline
        Ours                                         & \textbf{3.11} & 5.48  & \textbf{12.26} & \textbf{8.14}          & \textbf{3.18}  & \textbf{5.80}\\
        \hline
    \end{tabular}
\end{table*}

\begin{table*}
    \centering
    \caption{ Quantitative evaluation results of the compared methods and our method on the synthetic point clouds in Figures \ref{fig:bunnyhi}, \ref{fig:rockerman}, \ref{fig:Icosahedron}, \ref{fig:Dodecahedron} and \ref{fig:kitten}. Note that * represents deep learning methods. Mean Square Error ($\times10^{-3}$) is used here. The best method for each model is highlighted. }
    \label{table:errors_evaluation_mse_syn}
    \begin{tabular}{l c c c c c c}
        \hline
        Methods& {Figure \ref{fig:bunnyhi}}& {Figure \ref{fig:rockerman}}& {Figure \ref{fig:Icosahedron}}& {Figure \ref{fig:Dodecahedron}} & {Figure \ref{fig:kitten}}& Avg.\\ 
        \hline
        CLOP \cite{preiner2014continuous_CLOP}            & 10.32          & 13.91           & 21.86 & 23.31          & 9.88          & 15.86\\
        GPF \cite{lu2017gpf_GPF}                         & 11.64         & 17.07          & 22.43         & 23.88         & 11.97          & 17.40\\
        RIMLS \cite{oztireli2009feature_RIMLS}             & 10.05         & 14.02           & 21.68         & 23.30 & 10.02          & 15.81\\
        TD* \cite{Hermosilla_2019_ICCV_TotalDenoising}& 13.22 & 14.78  & 22.44          & 23.36 & 11.45  & 17.05\\
        PCN* \cite{rakotosaona2020pointcleannet_PointCleanNet} & 10.30 & 14.28  & 23.68          & 23.79 & 10.59  & 16.53\\
        PF* \cite{zhang2020pointfilter_Pointfilter}          & 10.02 & 14.17  & 23.71 & 25.28 & \textbf{9.82}  & 16.60\\
        \hline
        Ours                                         & \textbf{9.92} & \textbf{14.01}  & \textbf{21.46} & \textbf{23.15}          & 9.92  & \textbf{15.69}\\
        \hline
    \end{tabular}
\end{table*}

\subsection{Quantitative Comparisons}
We also make a quantitative comparison of the two evaluation metrics introduced in the previous section. 
Note that since there is no corresponding ground-truth model for the point clouds with raw scanned noise, we  choose the models with synthetic noise for quantitative evaluation. 
The results under the Chamfer Distance metric are given in Table \ref{table:errors_evaluation_chamfer_syn}. Despite the fact that deep learning-based methods are trained on a large number of point clouds, our method still outperforms all deep learning-based methods and even achieves the lowest quantitative results among most models. 
In terms of the other evaluation metric MSE, our method still outperforms most deep learning methods and holds the lowest quantitative error among most models, as shown in Table \ref{table:errors_evaluation_mse_syn}.

These quantitative results remain consistent with the visual results, demonstrating that our method generally outperforms existing methods both visually and quantitatively. 
We consider it is because our method provides a better uniform distribution of the filtered points and can handle both sparsely and densely sampled point clouds.
In the case of sparse sampling, some deep learning-based methods are unable to obtain meaningful local geometric information from the sparse points of local neighbors.
It is also worth noting that although RIMLS achieves comparable results to ours in some of the visual results, its error values are greater than those of our method in most cases due to its uneven point cloud distribution.

\subsection{Ablation Study}
\begin{figure}[!h]
\centering
\begin{minipage}[b]{0.32\linewidth}
\subfigure[Noisy input]{\label{fig:noise_input_1.0}\includegraphics[width=1\linewidth]{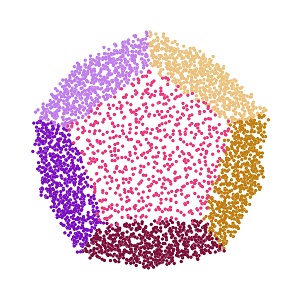}}
\end{minipage}
\begin{minipage}[b]{0.32\linewidth}
\subfigure[$k$ = 1]{\label{fig:knn_1}\includegraphics[width=1\linewidth]{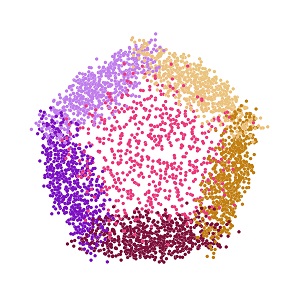}}
\end{minipage}
\begin{minipage}[b]{0.32\linewidth}
\subfigure[$k$ = 5]{\label{fig:knn_5}\includegraphics[width=1\linewidth]{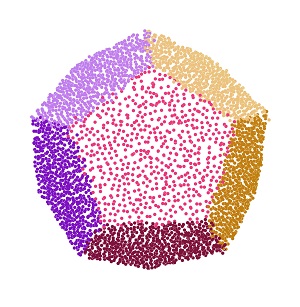}}
\end{minipage}
\begin{minipage}[b]{0.32\linewidth}
\subfigure[$k$ = 15]{\label{fig:knn_15}\includegraphics[width=1\linewidth]{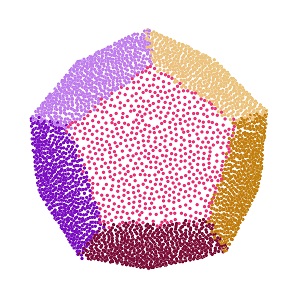}}
\end{minipage} 
\begin{minipage}[b]{0.32\linewidth}
\subfigure[$k$ = 30]{\label{fig:knn_30}\includegraphics[width=1\linewidth]{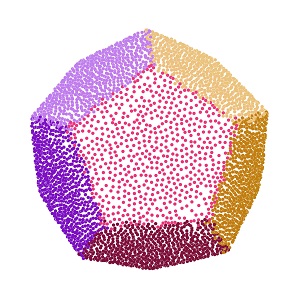}}
\end{minipage}
\begin{minipage}[b]{0.32\linewidth}
\subfigure[$k$ = 45]{\label{fig:knn_45}\includegraphics[width=1\linewidth]{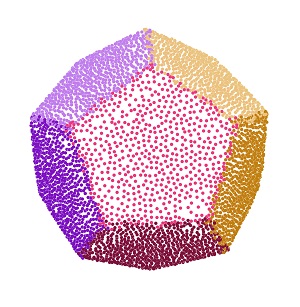}}
\end{minipage}
\\
\caption{Filtered results with different $k$. Noise level: 1.0$\%$. $t$ = 30, $\mu$ = 0.3. }
\label{fig:knn}
\end{figure}

\textbf{Parameters. }
We first perform experiments on a point cloud containing 7682 points with different values of $k$. From Figure \ref{fig:knn}, the best value of $k$ is 30. This is also the default value for this parameter. 
It is easy to know that the range of $k$ is highly related to the density of the point cloud. 
With a fixed value of $k$, when the model has a sparser distribution, the local range delimited by $k$ will become larger, which may lead to an excessive range that should not be treated as local information, resulting in a less desired outcome. For point clouds with denser distribution, the local neighbors' $k$ range becomes smaller, meaning that the $k$ neighborhood contains only a smaller range of local information, further leading to an uneven distribution of the point cloud. Generally, we use a larger $k$ for point clouds with denser points to ensure an appropriate number of local neighbors. 

\begin{figure}[!h]
\centering
\begin{minipage}[b]{0.32\linewidth}
\subfigure[Noisy input]{\label{fig:noise_input_1.0_miuview}\includegraphics[width=1\linewidth]{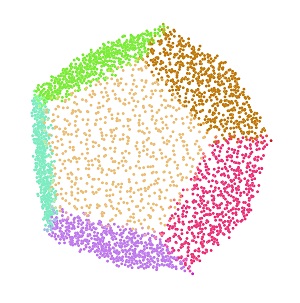}}
\end{minipage}
\begin{minipage}[b]{0.32\linewidth}
\subfigure[$\mu$ = 0.1]{\label{fig:miu_0.1}\includegraphics[width=1\linewidth]{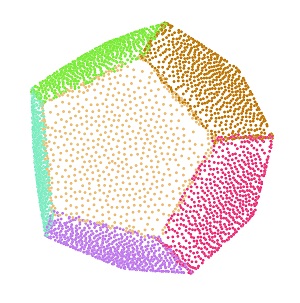}}
\end{minipage}
\begin{minipage}[b]{0.32\linewidth}
\subfigure[$\mu$ = 0.2]{\label{fig:miu_0.2}\includegraphics[width=1\linewidth]{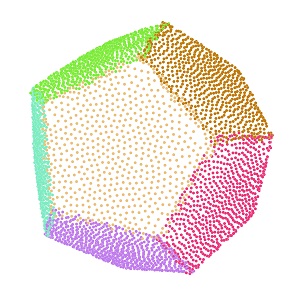}}
\end{minipage}
\begin{minipage}[b]{0.32\linewidth}
\subfigure[$\mu$ = 0.3]{\label{fig:miu_0.3}\includegraphics[width=1\linewidth]{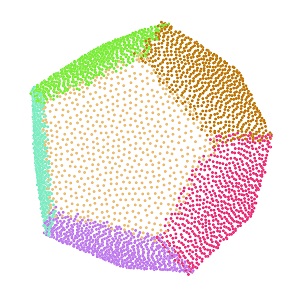}}
\end{minipage} 
\begin{minipage}[b]{0.32\linewidth}
\subfigure[$\mu$ = 0.4]{\label{fig:miu_0.4}\includegraphics[width=1\linewidth]{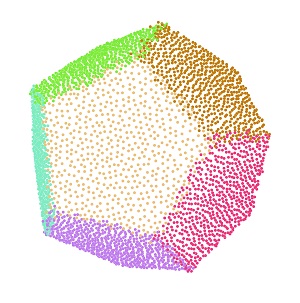}}
\end{minipage}
\begin{minipage}[b]{0.32\linewidth}
\subfigure[$\mu$ = 0.5]{\label{fig:miu_0.5}\includegraphics[width=1\linewidth]{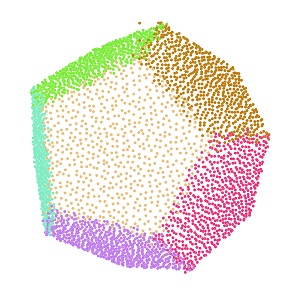}}
\end{minipage}
\\
\caption{Filtered results with different $\mu$. Noise level: 1.0$\%$, $t$ = 30, $k$ = 30. }
\label{fig:miu}
\end{figure}

As $\mu$ is related to the number of iterations $t$, we give
filtered results for different values of $\mu$ under certain iterations.
Figure \ref{fig:miu} demonstrates the filtered point clouds obtained for different $\mu$ values when $t$ = 30 and $k$ = 30. We can see from this figure that as $\mu$ increases, the distribution of the point cloud becomes more uniform, but a too-large $\mu$ will make the model turn into chaos again. 
Figure \ref{fig:miu_0.1} shows the filtered results with a low value of $\mu$ when $i$=30 and $k$=30. As we can see, a smaller $\mu$ is better for maintaining the feature edges of the model.

\begin{figure}[!h]
\centering
\begin{minipage}[b]{0.24\linewidth}
\subfigure[Noisy input]{\label{fig:noise_input_0.5}\includegraphics[width=1\linewidth]{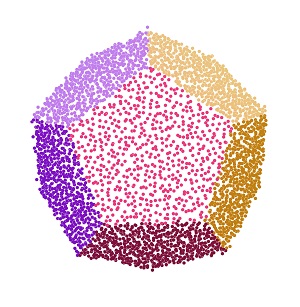}}
\end{minipage}
\begin{minipage}[b]{0.24\linewidth}
\subfigure[5 iterations]{\label{fig:iter_num_5}\includegraphics[width=1\linewidth]{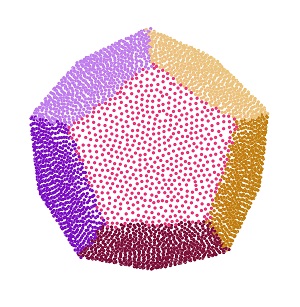}}
\end{minipage}
\begin{minipage}[b]{0.24\linewidth}
\subfigure[15 iterations]{\label{fig:iter_num_15}\includegraphics[width=1\linewidth]{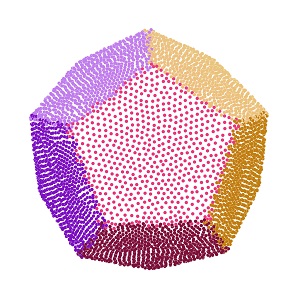}}
\end{minipage} 
\begin{minipage}[b]{0.24\linewidth}
\subfigure[30 iterations]{\label{fig:iter_num_30}\includegraphics[width=1\linewidth]{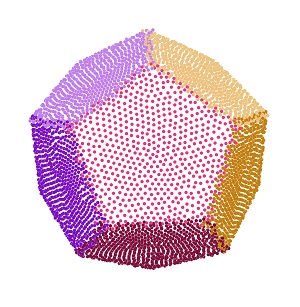}}
\end{minipage}
\caption{Filtered results of different iterations. Noise level: 1.0$\%$, other parameters: $k$ = 30, $\mu$ = 0.3. }
\label{fig:iter_num}
\end{figure}

We also conduct experiments under different iterations. Figure \ref{fig:iter_num} indicates that with an increasing number of iteration, the distribution of the filtered point cloud becomes more uniform. However, Figure \ref{fig:iter_num_30} shows that if the iteration parameter is set too large, the boundary of the model will become unclear again. 

\begin{figure}[!h]
\centering
\begin{minipage}[b]{0.35\linewidth}
\subfigure[Without repulsion term]{\label{fig:withoutRepulsion}\includegraphics[width=1\linewidth]{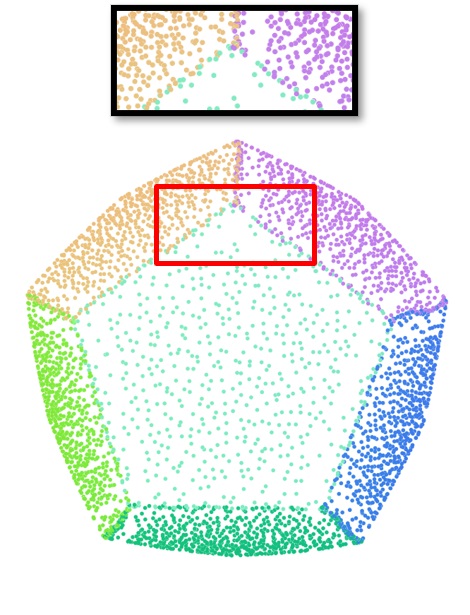}}
\end{minipage}
\begin{minipage}[b]{0.35\linewidth}
\subfigure[With repulsion term]{\label{fig:withRepulsion}\includegraphics[width=1\linewidth]{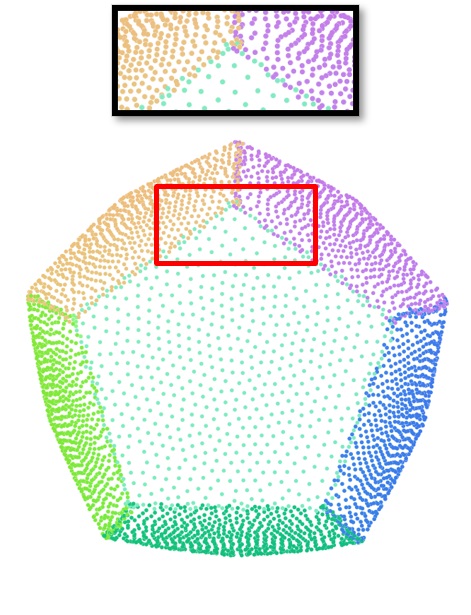}}
\end{minipage}
\\
\caption{Filtered results with or without the repulsion term. }
\label{fig:repulsion}
\end{figure}

\textbf{With/without the repulsion term. }
Local-based filtering approaches tend to converge in certain places when updating the positions. Obviously, this will make follow-up applications such as surface reconstruction very difficult. 
Our method adopts the repulsion term mentioned in Section \ref{sec:method} to allow the points to be evenly distributed while filtering, thus improving the quality of the filtered point cloud. 
As shown in Figure \ref{fig:withoutRepulsion}, without the repulsion term, it is clear that some points are concentrated at the edges, whereas the distribution in Figure \ref{fig:withRepulsion} is more even.

\begin{figure}[!h]
\centering
\begin{minipage}[c]{0.1\linewidth}
\begin{center}
Noisy input
\end{center}
\end{minipage}
\begin{minipage}[c]{0.28\linewidth}
{\label{}\includegraphics[width=1\linewidth]{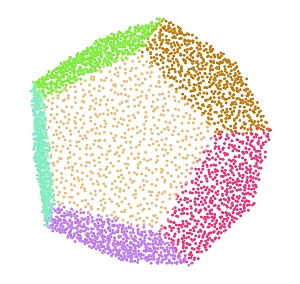}}
\end{minipage}
\begin{minipage}[c]{0.28\linewidth}
{\label{}\includegraphics[width=1\linewidth]{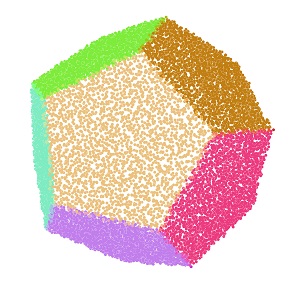}}
\end{minipage}
\begin{minipage}[c]{0.28\linewidth}
{\label{}\includegraphics[width=1\linewidth]{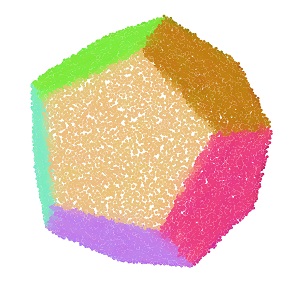}}
\end{minipage} 
\\
\begin{minipage}[c]{0.1\linewidth}
\begin{center}
Ours
\end{center}
\end{minipage}
\begin{minipage}[c]{0.28\linewidth}
{\label{}\includegraphics[width=1\linewidth]{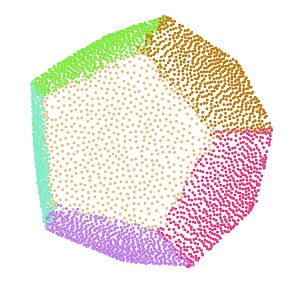}}
\end{minipage}
\begin{minipage}[c]{0.28\linewidth}
{\label{}\includegraphics[width=1\linewidth]{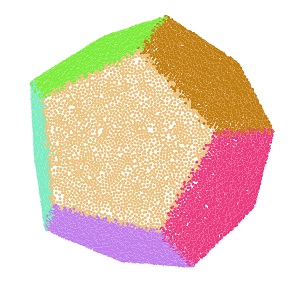}}
\end{minipage}
\begin{minipage}[c]{0.28\linewidth}
{\label{}\includegraphics[width=1\linewidth]{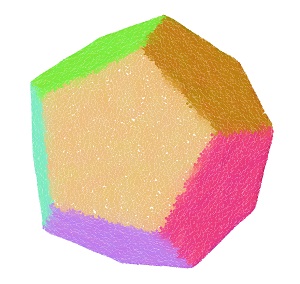}}
\end{minipage} 
\\
\begin{minipage}[c]{0.1\linewidth}
\begin{center}
\end{center}
\end{minipage}
\begin{minipage}[c]{0.28\linewidth}
\centerline{7682 points}
\end{minipage}
\begin{minipage}[c]{0.28\linewidth}
\centerline{30722 points}
\end{minipage}
\begin{minipage}[c]{0.28\linewidth}
\centerline{67938 points}
\end{minipage}
\caption{ Filtered results of models with different sampling numbers of points. }
\label{pointNumber}
\label{fig:pointnumber}
\end{figure}

\textbf{Point density. }
The performance under different point densities is tested. Figure \ref{fig:pointnumber} shows that our method yields promising results on both sparse and dense point clouds. It is worth noting that since our method requires only local information, for point clouds with greater point density, a desired filtered result can be obtained by setting a larger $k$.

\begin{figure*}[!h]
\centering
\begin{minipage}[c]{0.1\linewidth}
\begin{center}
Noisy input
\end{center}
\end{minipage}
\begin{minipage}[c]{0.14\linewidth}
{\label{}\includegraphics[width=1\linewidth]{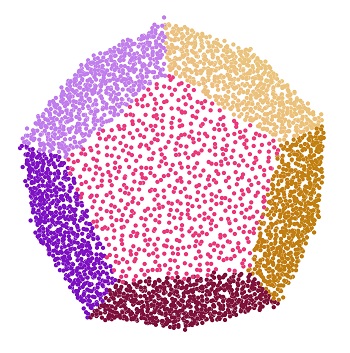}}
\end{minipage}
\begin{minipage}[c]{0.14\linewidth}
{\label{}\includegraphics[width=1\linewidth]{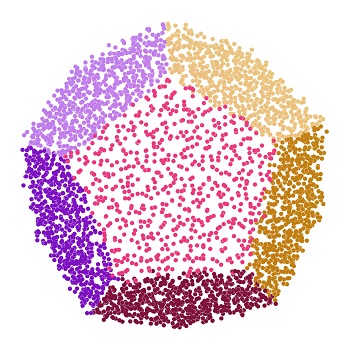}}
\end{minipage}
\begin{minipage}[c]{0.14\linewidth}
{\label{}\includegraphics[width=1\linewidth]{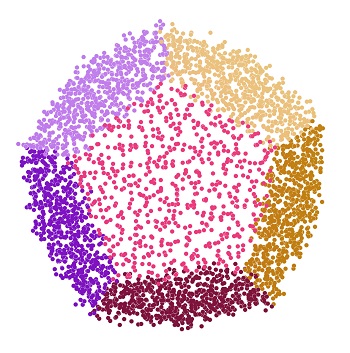}}
\end{minipage} 
\begin{minipage}[c]{0.14\linewidth}
{\label{}\includegraphics[width=1\linewidth]{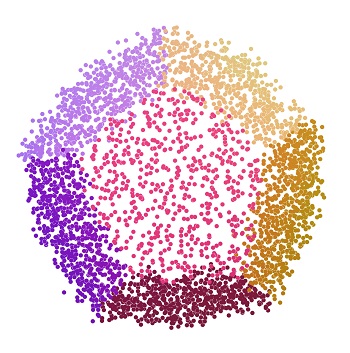}}
\end{minipage}
\begin{minipage}[c]{0.14\linewidth}
{\label{}\includegraphics[width=1\linewidth]{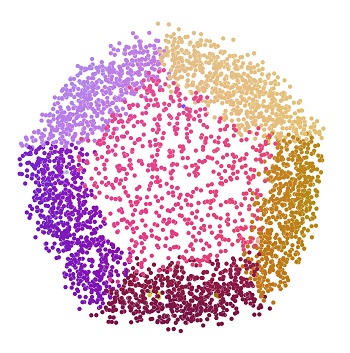}}
\end{minipage}
\begin{minipage}[c]{0.14\linewidth}
{\label{}\includegraphics[width=1\linewidth]{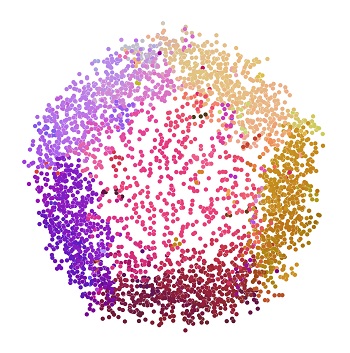}}
\end{minipage} 
\\
\begin{minipage}[c]{0.1\linewidth}
\begin{center}
Ours
\end{center}
\end{minipage}
\begin{minipage}[c]{0.14\linewidth}
{\label{}\includegraphics[width=1\linewidth]{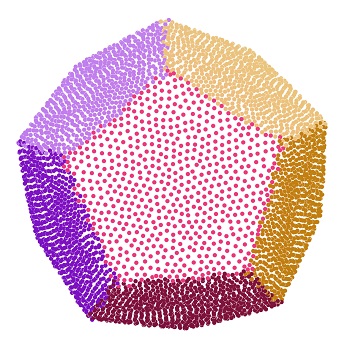}}
\end{minipage}
\begin{minipage}[c]{0.14\linewidth}
{\label{}\includegraphics[width=1\linewidth]{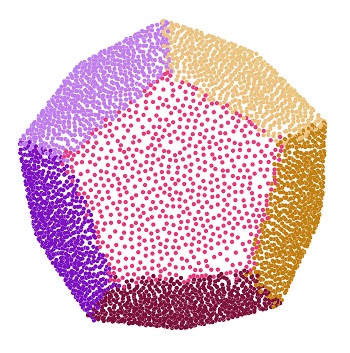}}
\end{minipage}
\begin{minipage}[c]{0.14\linewidth}
{\label{}\includegraphics[width=1\linewidth]{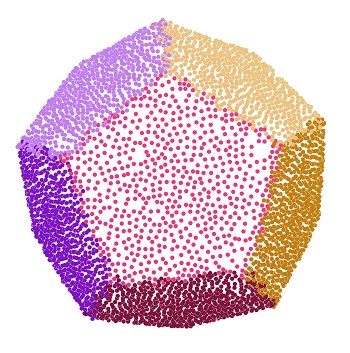}}
\end{minipage} 
\begin{minipage}[c]{0.14\linewidth}
{\label{}\includegraphics[width=1\linewidth]{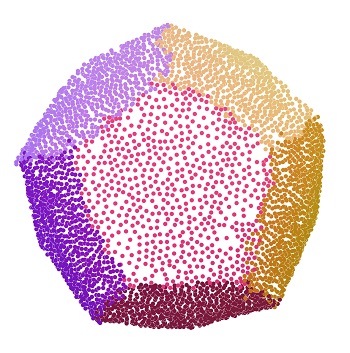}}
\end{minipage}
\begin{minipage}[c]{0.14\linewidth}
{\label{}\includegraphics[width=1\linewidth]{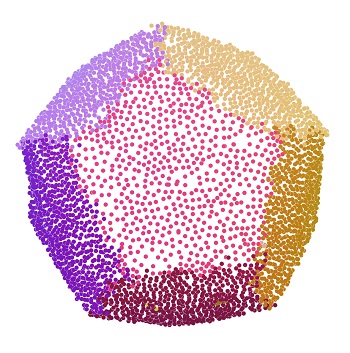}}
\end{minipage}
\begin{minipage}[c]{0.14\linewidth}
{\label{}\includegraphics[width=1\linewidth]{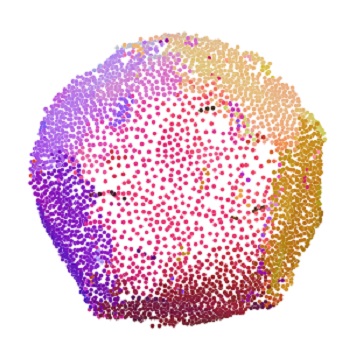}}
\end{minipage} 
\\
\begin{minipage}[c]{0.1\linewidth}
\begin{center}
\end{center}
\end{minipage}
\begin{minipage}[c]{0.14\linewidth}
\centerline{0.5\% noise}
\end{minipage}
\begin{minipage}[c]{0.14\linewidth}
\centerline{1.0\% noise}
\end{minipage}
\begin{minipage}[c]{0.14\linewidth}
\centerline{1.5\% noise}
\end{minipage} 
\begin{minipage}[c]{0.14\linewidth}
\centerline{2.0\% noise}
\end{minipage}
\begin{minipage}[c]{0.14\linewidth}
\centerline{2.5\% noise}
\end{minipage}
\begin{minipage}[c]{0.14\linewidth}
\centerline{3.0\% noise}
\end{minipage} 
\caption{Filtered results of models with different levels of noise.}
\label{fig:noiselv}
\end{figure*}

\textbf{Noise level. }
Different noise levels are applied to the same model to verify the robustness of our approach. 
Figure \ref{fig:noiselv} gives the filtered results by our method under noise levels of 0.5\%, 1.0\%, 1.5\%, 2.0\%, 2.5\% and 3.0\%. It can be seen that our method is capable of handling models with different levels of noise but may produce less desired results on excessively high noise. 
As our method relies on the quality of normals, it is difficult to accurately keep the geometric features if the model has less accurate normals caused by higher noise levels.

\begin{figure}[!h]
\centering
\begin{minipage}[b]{0.32\linewidth}
{\label{}\includegraphics[width=1\linewidth]{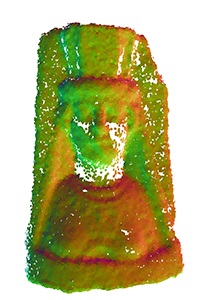}}
\centerline{PCN \cite{rakotosaona2020pointcleannet_PointCleanNet}}
\end{minipage}
\begin{minipage}[b]{0.32\linewidth}
{\label{}\includegraphics[width=1\linewidth]{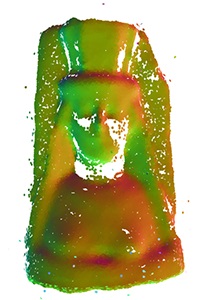}}
\centerline{PF \cite{zhang2020pointfilter_Pointfilter}}
\end{minipage}
\begin{minipage}[b]{0.32\linewidth}
{\label{}\includegraphics[width=1\linewidth]{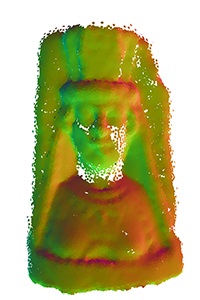}}
\centerline{Ours}
\end{minipage} 
\caption{Filtered results on the irregularly sampled point cloud.}
\label{fig:irrre_sampled}
\end{figure}

\textbf{Irregular sampling. }
We conduct experiments on models with irregular sampling. 
As shown in Figure \ref{fig:irrre_sampled}, we provide the visual comparisons on an unevenly sampled model for PCN \cite{rakotosaona2020pointcleannet_PointCleanNet}, PF \cite{zhang2020pointfilter_Pointfilter}, and our method. It can be seen that the filtered point cloud of PCN still contains obvious noise and PF makes the detail features blurred, while our method smooths the model better with feature preserving effect.

\begin{figure}[!h]
\centering
\begin{minipage}[b]{0.24\linewidth}
\subfigure[Cube with big holes]{\label{fig:noise_input_0.5}\includegraphics[width=1\linewidth]{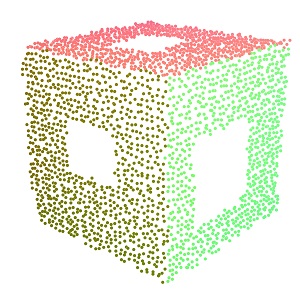}}
\end{minipage}
\begin{minipage}[b]{0.24\linewidth}
\subfigure[Cube with small holes]{\label{fig:iter_num_5}\includegraphics[width=1\linewidth]{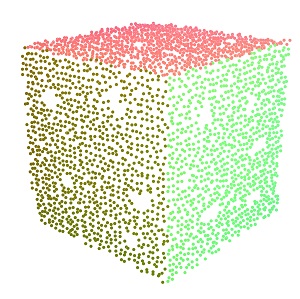}}
\end{minipage}
\begin{minipage}[b]{0.24\linewidth}
\subfigure[Updated points of (a)]{\label{fig:iter_num_15}\includegraphics[width=1\linewidth]{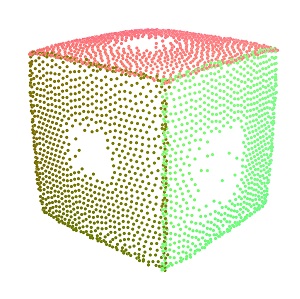}}
\end{minipage} 
\begin{minipage}[b]{0.24\linewidth}
\subfigure[Updated points of (b)]{\label{fig:iter_num_30}\includegraphics[width=1\linewidth]{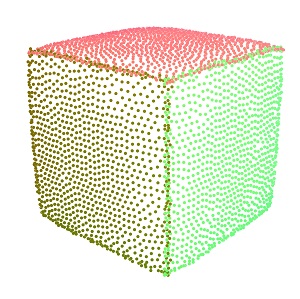}}
\end{minipage}
\caption{Filtered results of the model with holes.}
\label{fig:fillingholes}
\end{figure}

\textbf{Holes filling. }
Taking the cube as an example, we experiment on a model with holes. Figure \ref{fig:fillingholes} shows the filtered results of different holes. Our method is capable of filling relatively small holes because we consider the distribution of the updated points. However, it will be challenging to fill big holes that severely disrupt the surfaces of the model.

\begin{figure}[!h]
\centering
\begin{minipage}[b]{0.32\linewidth}
\subfigure[Noisy input]{\label{fig:lowRankRaw}\includegraphics[width=1\linewidth]{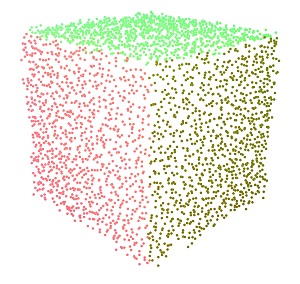}}
\end{minipage}
\begin{minipage}[b]{0.32\linewidth}
\subfigure[Lowrank\cite{lu2020low_Lowrank}]{\label{fig:lowRankResult}\includegraphics[width=1\linewidth]{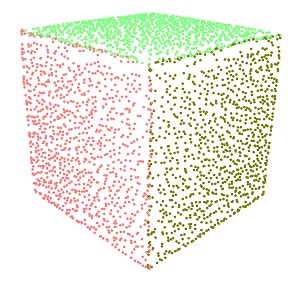}}
\end{minipage}
\begin{minipage}[b]{0.32\linewidth}
\subfigure[Ours]{\label{fig:lowRankOurs}\includegraphics[width=1\linewidth]{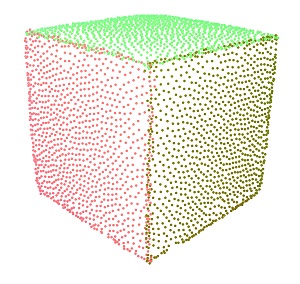}}
\end{minipage}
\\
\caption{Filtered points of ours and Lowrank \cite{lu2020low_Lowrank}.}
\label{fig:lowRank}
\end{figure}

\textbf{Lowrank \cite{lu2020low_Lowrank} versus ours. }
Figure \ref{fig:lowRank} gives a comparison of our approach with Lowrank \cite{lu2020low_Lowrank}.  
It demonstrates that our method gets a more uniform point distribution while removing noise, which has better quality than Lowrank.

\begin{figure}[htb]
\centering
\begin{minipage}[b]{0.30\linewidth}
\subfigure[Noisy input]{\label{fig:joint_raw}\includegraphics[width=1\linewidth]{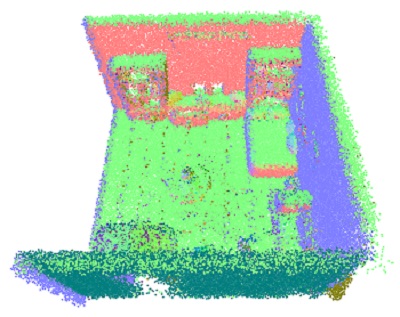}}
\end{minipage}
\begin{minipage}[b]{0.30\linewidth}
\subfigure[Filtered result]{\label{fig:joint_updated}\includegraphics[width=1\linewidth]{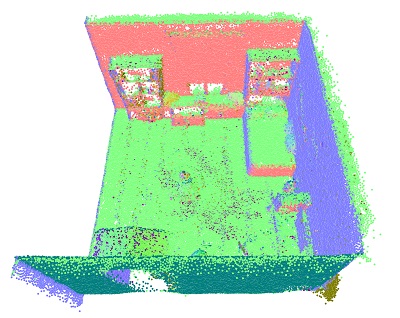}}
\end{minipage}
\\
\caption{Filtered result on noisy point cloud of an indoor scene. }
\label{fig:indoor}
\end{figure}

\textbf{Indoor scene data.}
We perform an experiment on the more challenging indoor scene data, as shown in Figure \ref{fig:indoor}. The result manifests that our method has the capacity of dealing with point cloud indoor scenes as well. 

\textbf{Runtime.}
The running time of the proposed method is calculated under different $k$ and iterations parameters.  
It is clearly shown in Table \ref{table:runtime_hexadron} that as $k$ and the number of iterations increase, the runtime increases accordingly. 

For each iteration, our method gathers $k$-local neighbors for each point to obtain the updated points. Thus the larger the parameter $k$ is, the slower its computation becomes. The number of iterations has a similar effect on the running time. The higher the number of iteration is, the longer the computation becomes. However, since our method is locally based, multiple iterations usually run fast.

In addition, we also give the runtime of other methods for comparison. Table \ref{table:runtime_compare} shows that our method is significantly faster than the other methods. 

\begin{table}[!h]
\centering
\caption{Runtime (in seconds) on Dodecahedron for different $k$ and $t$. }
\begin{tabular}{l l l l l l}
\hline
Iterations     & $t$ = 5  & $t$ = 15 & $t$ = 30 & $t$ = 60          \\
\hline
$k$ = 30 & 1.41          & 3.77          & 7.37          & 14.69                  \\
$k$ = 60 & 2.33          & 6.37          & 12.36         & 24.36                        \\
\hline
\end{tabular}
\label{table:runtime_hexadron}
\end{table}

\begin{table}[!h]
\centering
\caption{Runtime (in seconds) comparison on different models. }
\begin{tabular}{p{1.1cm} p{1cm} p{1cm} p{1cm} p{1cm} p{1cm}}
\hline
Methods     & CLOP  & TD & PCN & PF & Ours        \\
\hline
Fig. \ref{fig:bunnyhi}      & 59.66    & 16.65    & 294.00    & 67.38 & \textbf{6.27} \\

Fig. \ref{fig:rockerman}  & 10.82   & 6.17   & 78.04     & 13.86  &\textbf{1.80}    \\

Fig. \ref{fig:Icosahedron}   & 3.89   & 4.91   & 83.69    & 38.98 &\textbf{1.89}    \\

Fig. \ref{fig:Dodecahedron} &2.60 &5.53 &28.24 &70.81 &\textbf{0.91} \\

Fig. \ref{fig:kitten}     & 42.85    & 16.24   & 186.30     & 49.99    &\textbf{4.66}  \\

Fig. \ref{fig:Nefertiti} & 60.92 & 155.76 & 317.67 & 81.10 & \textbf{7.93}  \\

Fig. \ref{fig:BuddhaStele} & 70.58 & 102.41 & 642.01 & 247.77 & \textbf{6.37}  \\

Fig. \ref{fig:Realscan} & 103.91 & 40.52 & 241.05 & 68.23 & \textbf{28.16}  \\

Fig. \ref{fig:david} & 50.58 & 30.01 & 352.99 & 74.68 & \textbf{6.98}  \\
\hline
\end{tabular}
\label{table:runtime_compare}
\end{table}

\begin{figure}[htb]
\centering
\begin{minipage}[b]{0.35\linewidth}
\subfigure[Noisy input]{\label{fig:joint_raw}\includegraphics[width=1\linewidth]{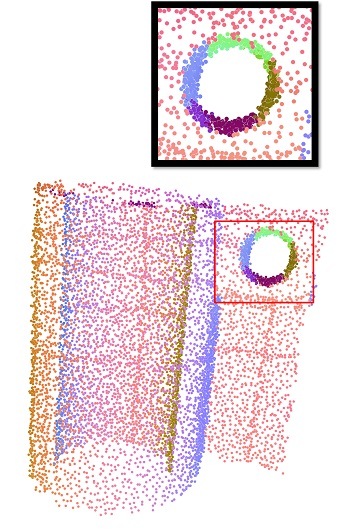}}
\end{minipage}
\begin{minipage}[b]{0.35\linewidth}
\subfigure[Filtered result]{\label{fig:joint_updated}\includegraphics[width=1\linewidth]{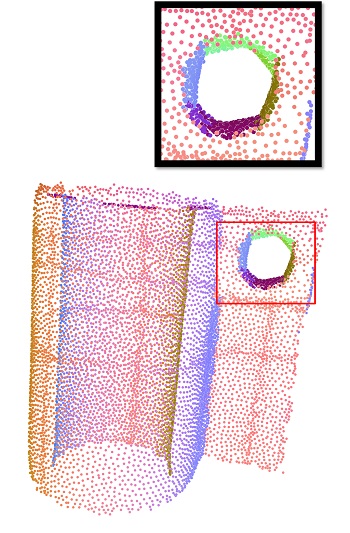}}
\end{minipage}
\\
\caption{A failure example.}
\label{fig:limitation}
\end{figure}

\textbf{Limitation.}
Though our method achieves good results, it still has room for improvement. Similar to \cite{lu2017gpf_GPF}, since it is a normal-based approach, it is inevitably dependent on the normal quality. In each iteration of the position update, each point is estimated with reference to the direction of the normal. Therefore, less accurate input normals may affect the filtered results. Figure \ref{fig:limitation} shows an example of this issue. Also, similar to previous methods, our method may produce less desirable results when handling a very high level of noise. For instance, Figure \ref{fig:noiselv} indicates 1.5\% noise is more challenging than the 0.5\% and 1.0\% noise. In future, we would like to develop effective techniques to handle the above limitations, e.g., fusing evolutionary optimization within the filtering framework  \cite{nakane2020application}.

%% file: paper/conclusion.tex
\section{Conclusion}
\label{sec:conclusion}
In this paper, we presented a method to improve point cloud filtering by enabling a more even point distribution for filtered point clouds. Built on top of \cite{lu2020low_Lowrank}, our method introduces a repulsion term into the objective function.
It not only removes noise while preserving sharp features but also ensures a more uniform distribution of cleaned points. Experiments show that our method obtains very promising filtered results under different levels of noise and densities. Both visual and quantitative comparisons also show that it generally outperforms the existing techniques in terms of visual quality and quantity. Our method also runs fast, exceeding the other compared methods. 